%% file: bare_jrnl_new_sample4.tex
\documentclass[lettersize,journal]{IEEEtran}
\usepackage{amsmath,amsfonts}
\usepackage{algorithmic}
\usepackage{algorithm}
\usepackage{array}
\usepackage[caption=false,font=normalsize,labelfont=sf,textfont=sf]{subfig}
\usepackage{textcomp}
\usepackage{stfloats}
\usepackage{url}
\usepackage{verbatim}
\usepackage{graphicx}
\usepackage{cite}
\usepackage{multirow}
\usepackage{makecell}
\usepackage{bbding}
\usepackage{booktabs}
\usepackage{color}
\usepackage{diagbox}

\begin{document}

\title{LVOS: A Benchmark for Large-scale \\ Long-term Video Object Segmentation}

\author{Lingyi Hong, Zhongying Liu, Wenchao Chen, Chenzhi Tan, Yuang Feng, Xinyu Zhou, Pinxue Guo, Jinglun Li, Zhaoyu Chen, Shuyong Gao, Wei Zhang, Wenqiang Zhang      
\thanks{Lingyi Hong, Zhongying Liu, Wenchao Chen, Chenzhi Tan, Yuang Feng, Xinyu Zhou, Shuyong Gao, Wei Zhang are with Shanghai Key Laboratory of Intelligent Information Processing, School of Computer Science, Fudan University, Shanghai 200433, China. E-mail: \{lyhong22, 21210240266, 21210240129, cztan23, yafeng23\}@m.fudan.edu.cn; \{zhouxinyu20, sygao18, weizh\}@fudan.edu.cn}
\thanks{Pinxue Guo, Jinglun Li, Zhaoyu Chen are with the Shanghai Engineering Research Center of AI\&Robotics, Academy for Engineering\&Technology, Fudan University, Shanghai, China. E-mail: \{pxguo21, jinglunli21\}@m.fudan.edu.cn; zhaoyuchen20@fudan.edu.cn}
\thanks{Wenqiang Zhang is with Engineering Research Center of AI\&Robotics, Ministry of Education, Academy for Engineering\&Technology, Fudan University, Shanghai, China, and also with the Shanghai Key Lab of Intelligent Information Processing, School of Computer Science, Fudan University, Shanghai, China. E-mail: wqzhang@fudan.edu.cn}
\thanks{(Corresponding author: Wei Zhang and Wenqiang Zhang.)}}

\markboth{Journal of \LaTeX\ Class Files,~Vol.~18, No.~9, September~2020}%
{Shell \MakeLowercase{\textit{et al.}}: A Sample Article Using IEEEtran.cls for IEEE Journals}


\maketitle

\input{./section/0_abstract}
\input{./section/1_intro}

\input{./section/2_related}

\input{./section/3_method}

\input{./section/4_experiment}

\input{./section/5_future_and_conclusion}

\section*{Acknowledgments}
This work was supported by National Natural Science Foundation of China (No.62072112), Scientific and Technological Innovation Action Plan of  Shanghai Science and Technology Committee (No.22511102202).

%


%

%

\bibliographystyle{IEEEtran}
\bibliography{./egbib}

\begin{IEEEbiography}[{\includegraphics[width=1in,height=1.25in,clip,keepaspectratio]{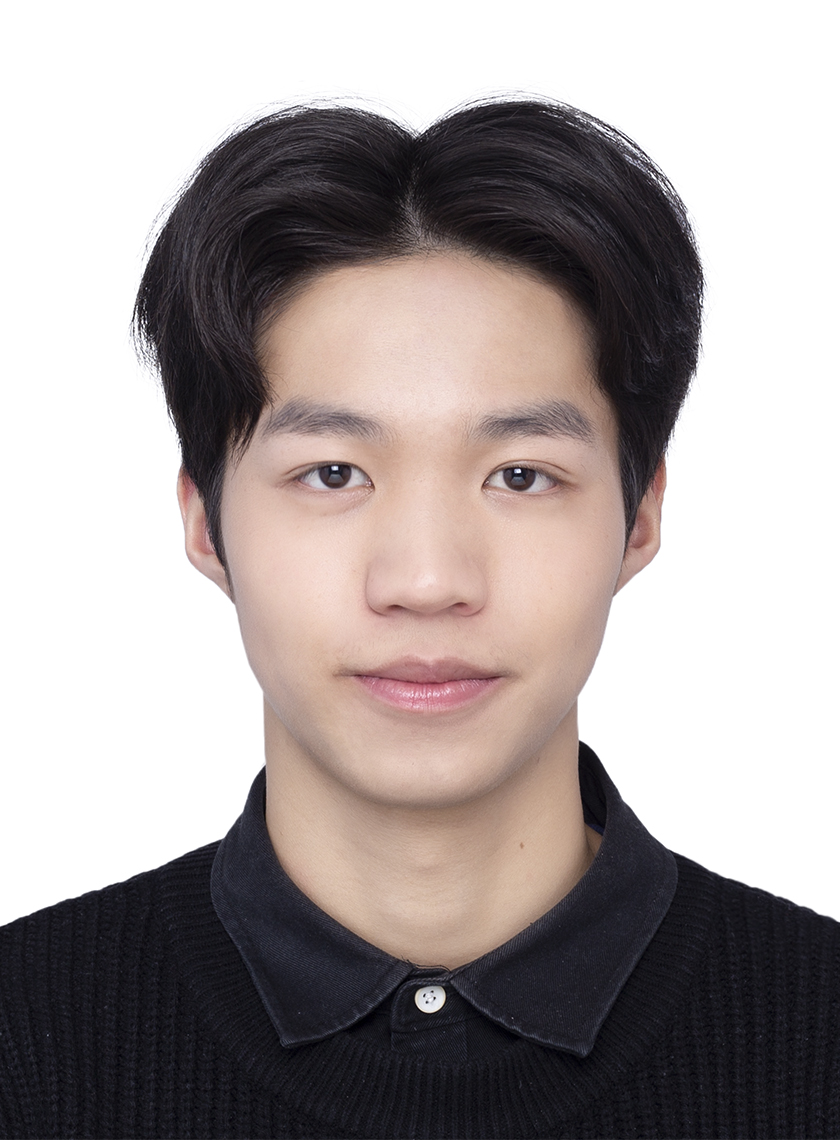}}]{Lingyi Hong}
is currently pursuing the Ph.D degree with the School of Computer Science, Fudan University, Shanghai, China. He received his B.S. degree in information security from Fudan University in 2022. His research interests include computer vision, video understanding, multimodal learning, and 3D vision.
\end{IEEEbiography}

\begin{IEEEbiography}[{\includegraphics[width=1in,height=1.25in,clip,keepaspectratio]{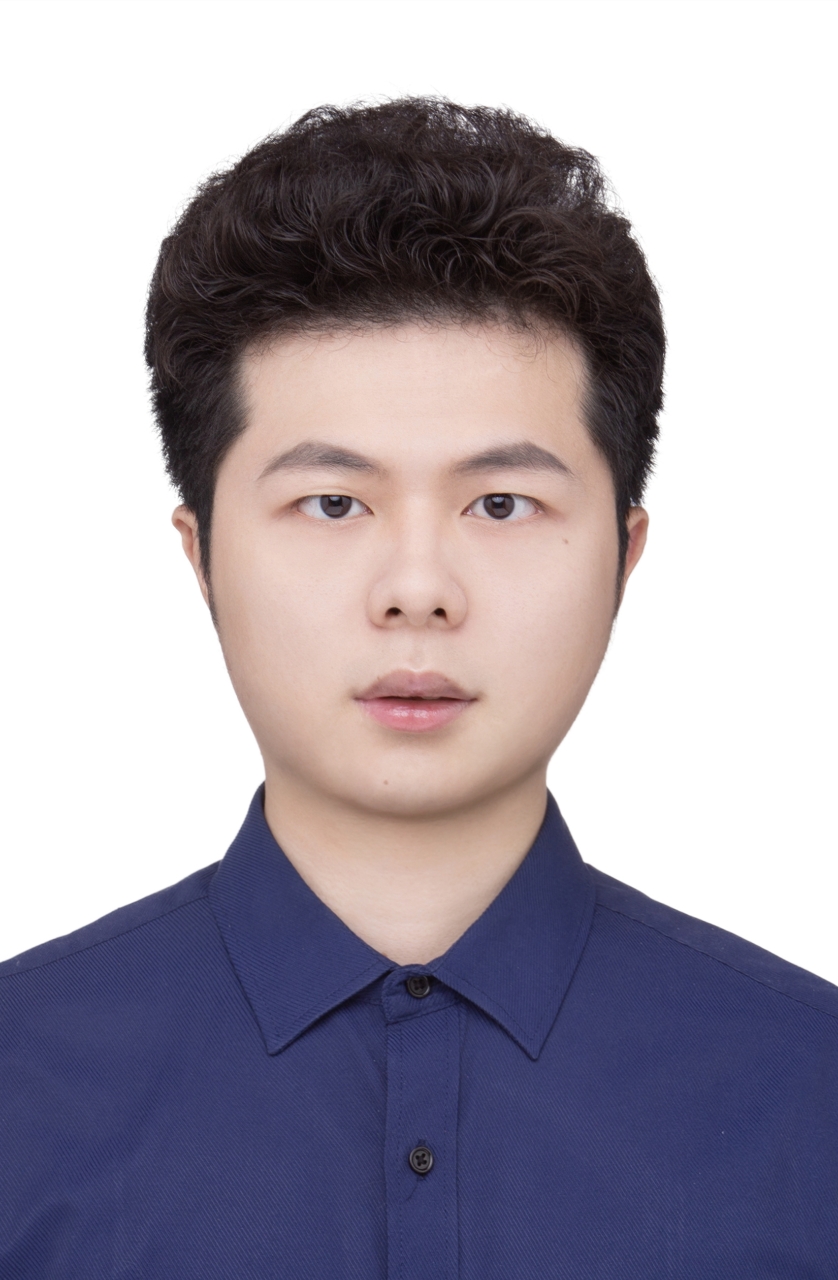}}]{Zhongying Liu}
obtained his master's degree from Fudan University in 2024. His research interests include computer vision, large language models, and multimodal learning.
\end{IEEEbiography}

\begin{IEEEbiography}[{\includegraphics[width=1in,height=1.25in,clip,keepaspectratio]{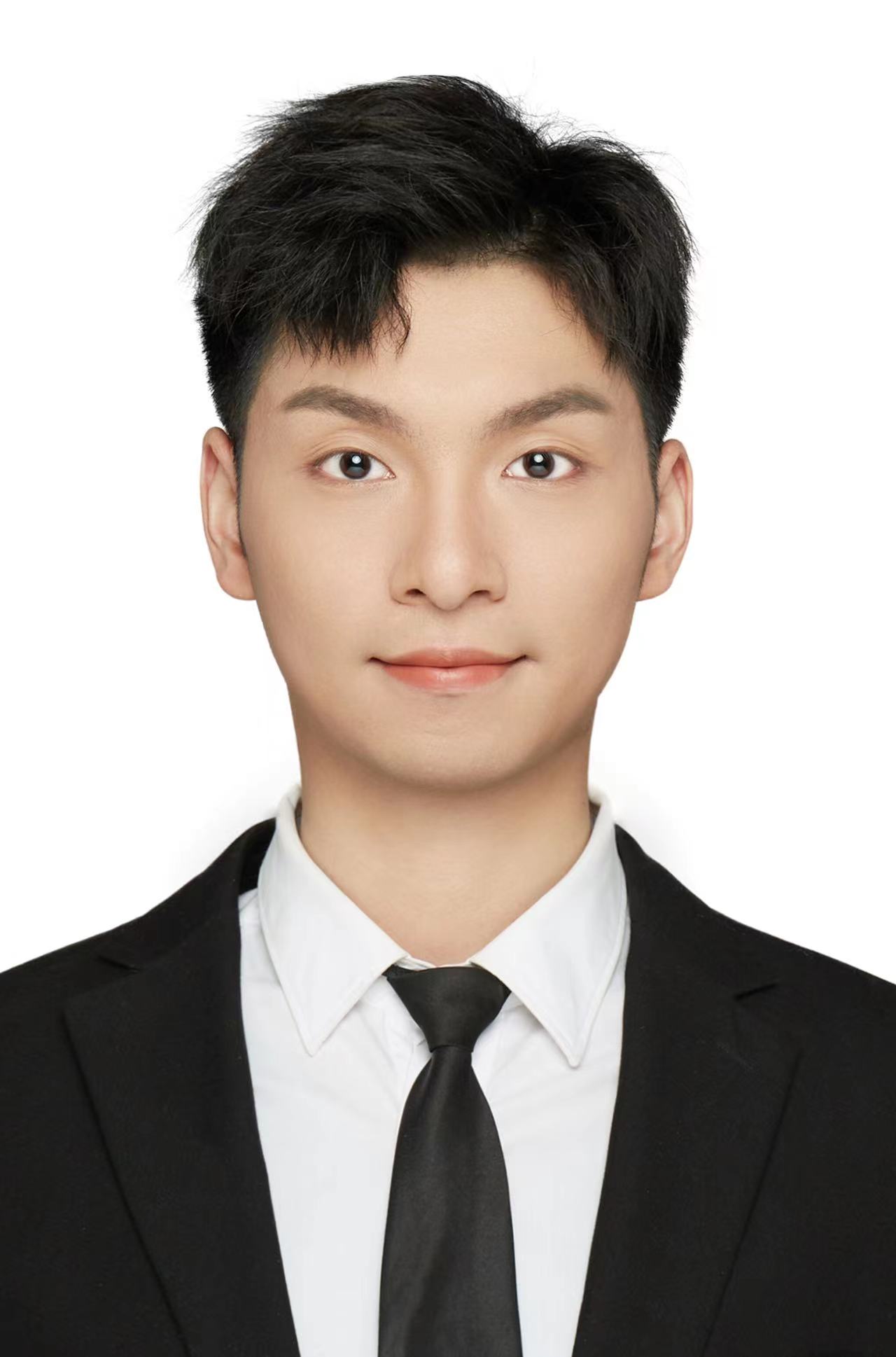}}]{Wenchao Chen}
is currently pursuing the Master degree with the School of Computer Science, Fudan University, Shanghai, China. His research interests include computer vision, video object segmentation, multimodal large language models.
\end{IEEEbiography}

\begin{IEEEbiography}[{\includegraphics[width=1in,height=1.25in,clip,keepaspectratio]{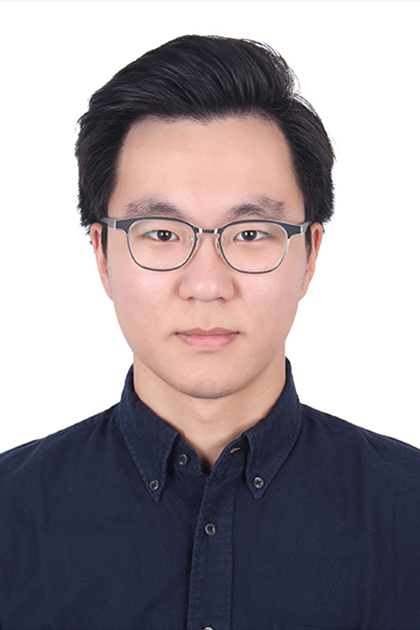}}]{Chenzhi Tan}
is currently pursuing the Master degree with the School of Computer Science, Fudan University, Shanghai, China. His research interests include robotic manipulation and perception.
\end{IEEEbiography}

\begin{IEEEbiography}[{\includegraphics[width=1in,height=1.25in,clip,keepaspectratio]{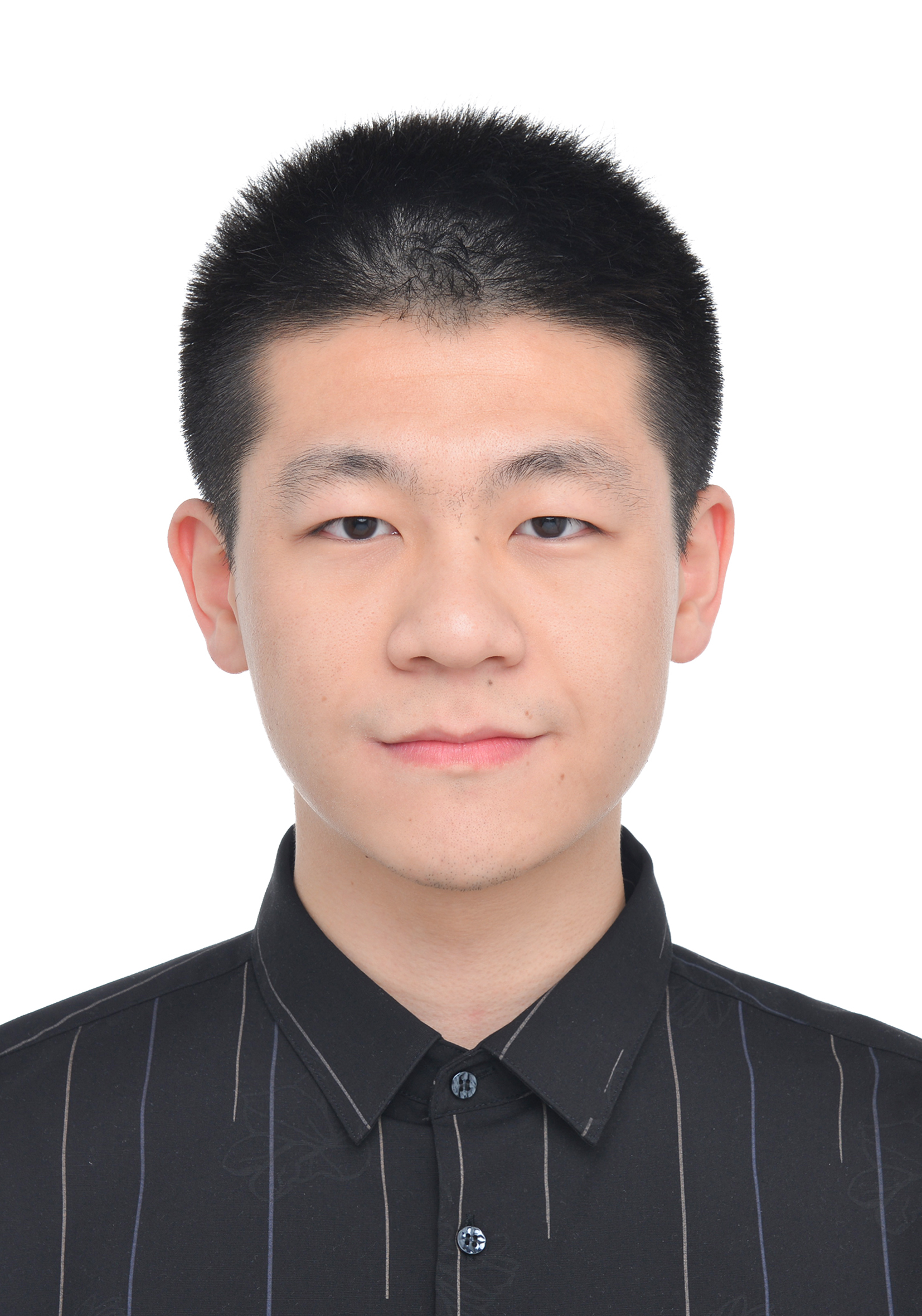}}]{Yuang Feng}
is currently pursuing the Master degree with the School of Computer Science, Fudan University, Shanghai, China. His research interests include computer vision, camouflaged object detection.
\end{IEEEbiography}

\begin{IEEEbiography}[{\includegraphics[width=1in,height=1.25in,clip,keepaspectratio]{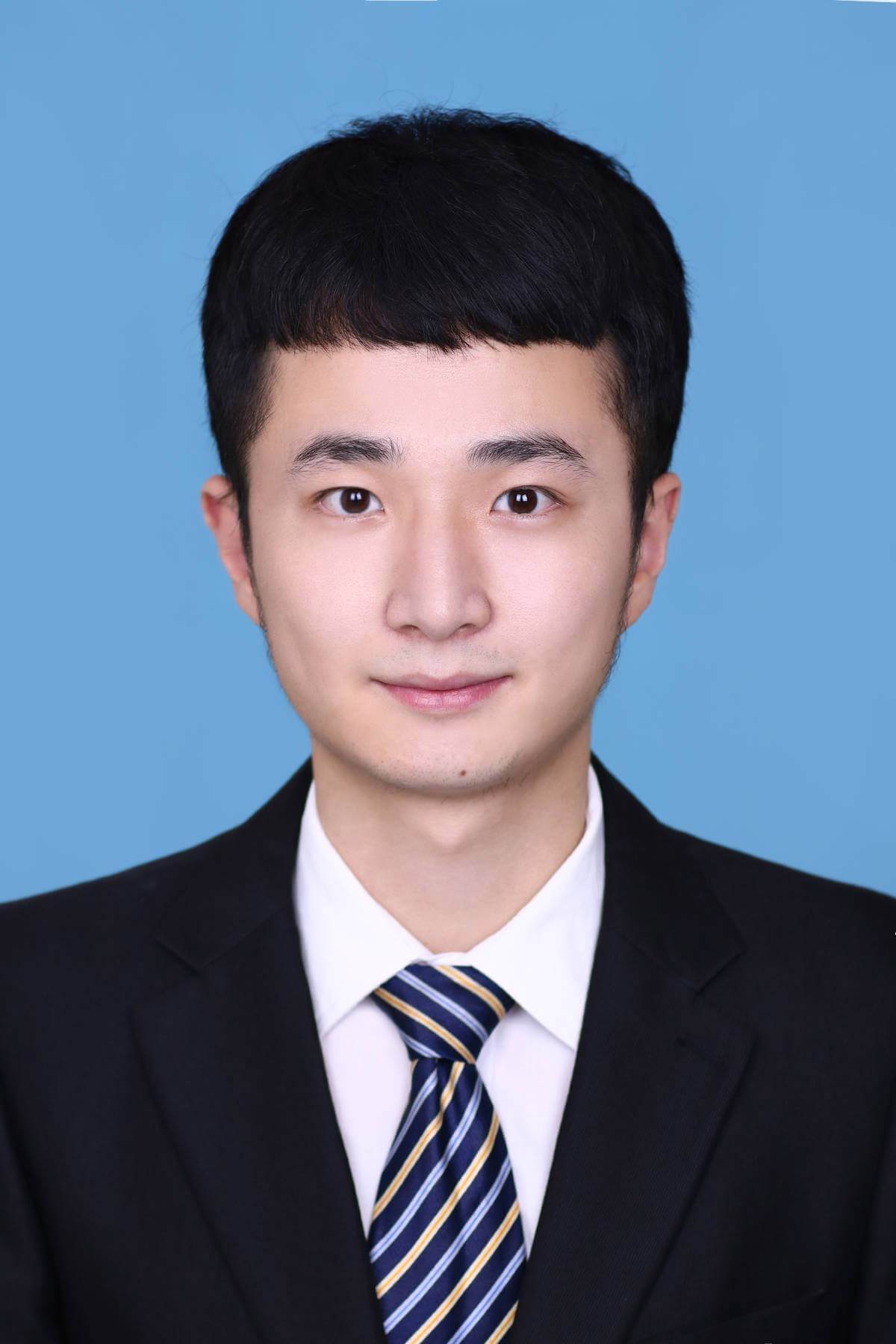}}]{Xinyu Zhou}
received the M.Eng. degree from Xiangtan University in 2020. He is currently pursuing Eng.D. degree with the School of Computer Science, Fudan University, Shanghai. His research interests include computer vision, visual object tracking, and semantic segmentation, medical image processing.
\end{IEEEbiography}

\begin{IEEEbiography}[{\includegraphics[width=1in,height=1.25in,clip,keepaspectratio]{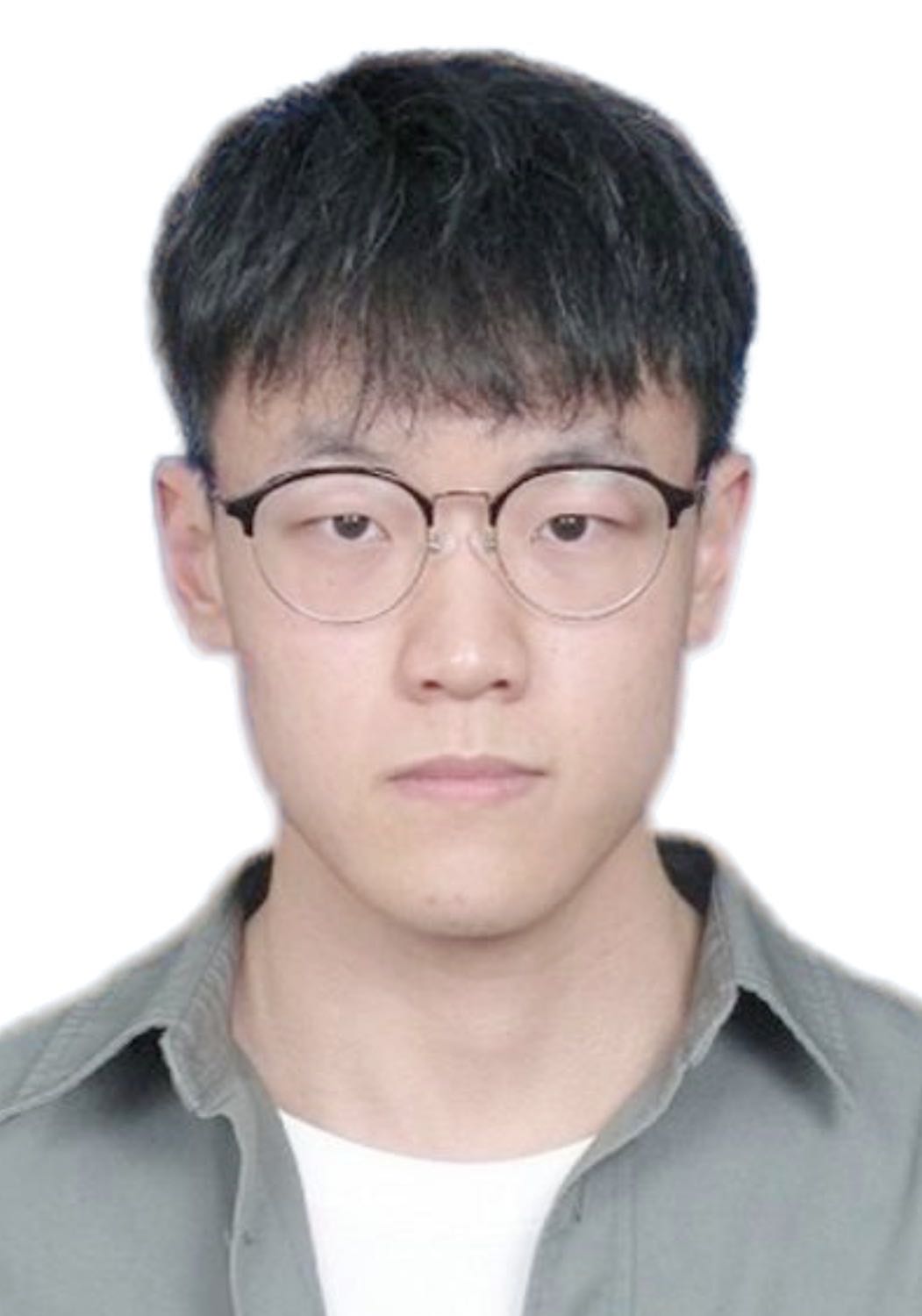}}]{Pinxue Guo}
received the B.S. degree in intelligence science and technology from Shanghai University in 2021. He is currently pursuing the Ph.D. degree with the Academy for Engineering and Technology, Fudan University, Shanghai, China. His research interests include video object segmentation, video understanding, and computer vision.
\end{IEEEbiography}

\begin{IEEEbiography}[{\includegraphics[width=1in,height=1.25in,clip,keepaspectratio]{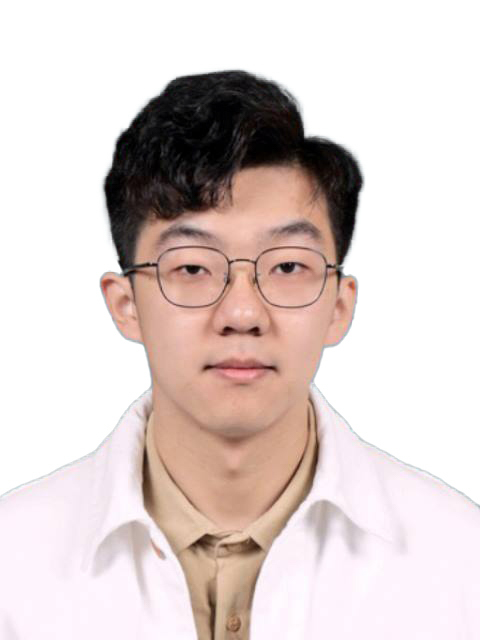}}]{Jinglun Li}
obtained a master's degree in Software Engineering in 2021 from Beijing University of Posts and Telecommunications. He is recently pursuing an Eng.D. in Computer Technology and Engineering at Fudan University. His current research interests include OOD detection, image segmentation, AIGC, and other areas within deep learning and machine learning.
\end{IEEEbiography}

\begin{IEEEbiography}[{\includegraphics[width=1in,height=1.25in,clip,keepaspectratio]{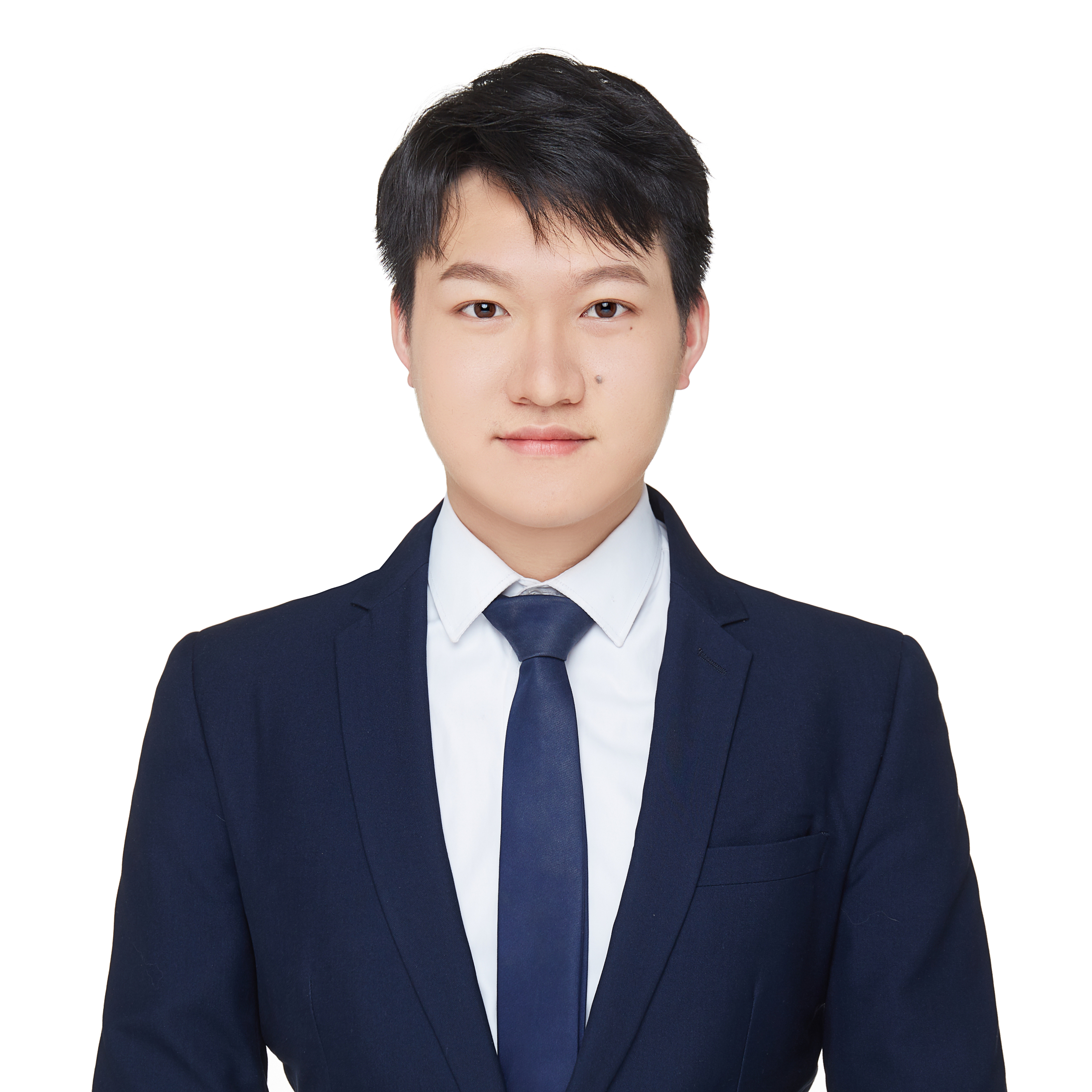}}]{Zhaoyu Chen}
received B.Eng. degree from Shandong University by 2020. He is currently pursuing the Ph.D. degree with the Academy for Engineering and Technology, Fudan University, Shanghai. His research interests include artificial intelligence security, computer vision, and their applications, such as adversarial examples and semantic segmentation.
\end{IEEEbiography}

\begin{IEEEbiography}[{\includegraphics[width=1in,height=1.25in,clip,keepaspectratio]{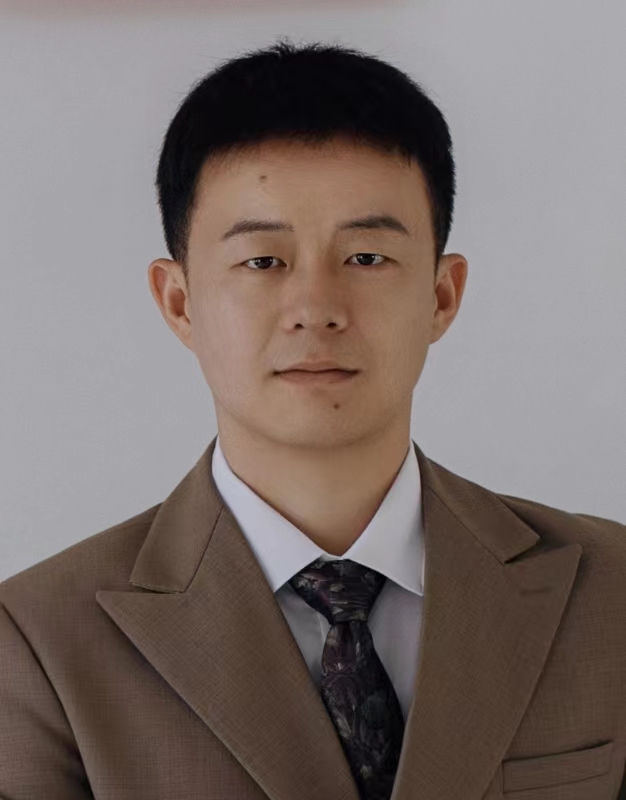}}]{Shuyong Gao}
is currently a postdoctoral researcher at Fudan University, Shanghai, China. His research interests include robotics, computer vision, generative models, and salient/camouflage object detection.
\end{IEEEbiography}

\begin{IEEEbiography}[{\includegraphics[width=1in,height=1.25in,clip,keepaspectratio]{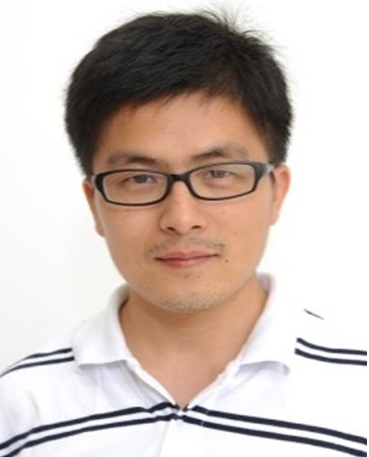}}]{Wei Zhang}
received the Ph.D. degree in computer science from Fudan University, China, in 2008. He was a Visiting Scholar at the UNC-Charlotte from 2016 to 2017. He is currently an Associate Professor with the School of Computer Science, Fudan University. His current research interests include deep learning, computer vision, and video object segmentation.
\end{IEEEbiography}

\begin{IEEEbiography}[{\includegraphics[width=1in,height=1.25in,clip,keepaspectratio]{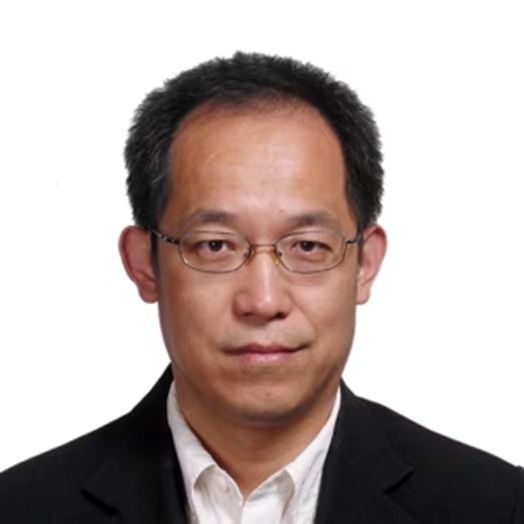}}]{Wenqiang Zhang}
is a professor with the School of Computer Science, Fudan University. He received his Ph.D. degree in Mechanical Engineering from Shanghai Jiao Tong University, China, in 2004. His current research interests include computer vision, and robot intelligence.
\end{IEEEbiography}

\end{document}

%% file: section/0_abstract.tex
\begin{abstract}
Video object segmentation (VOS) aims to distinguish and track target objects in a video. Despite the excellent performance achieved by off-the-shell VOS models, existing VOS benchmarks mainly focus on short-term videos lasting about 5 seconds, where objects remain visible most of the time. However, these benchmarks poorly represent practical applications, and the absence of long-term datasets restricts further investigation of VOS in realistic scenarios. Thus, we propose a novel benchmark named \textbf{LVOS}, comprising 720 videos with 296,401 frames and 407,945 high-quality annotations. Videos in LVOS last 1.14 minutes on average, approximately 5 times longer than videos in existing datasets. Each video includes various attributes, especially challenges deriving from the wild, such as long-term reappearing and cross-temporal similar objects. Compared to previous benchmarks, our LVOS better reflects VOS models' performance in real scenarios. Based on LVOS, we evaluate 20 existing VOS models under 4 different settings and conduct a comprehensive analysis. On LVOS, these models suffer a large performance drop, highlighting the challenge of achieving precise tracking and segmentation in real-world scenarios. Attribute-based analysis indicates that key factor to accuracy decline is the increased video length, emphasizing LVOS's crucial role. We hope our LVOS can advance development of VOS in real scenes. Data and code are available at \url{https://lingyihongfd.github.io/lvos.github.io/}.
\end{abstract}

\begin{IEEEkeywords}
Video Object Segmentation, Large-scale Benchmark, Long-term Video Understanding, Dataset.
\end{IEEEkeywords} 

%% file: section/1_intro.tex
\begin{figure*}[htbp]
    \centering
    \includegraphics[width=1.0\linewidth]{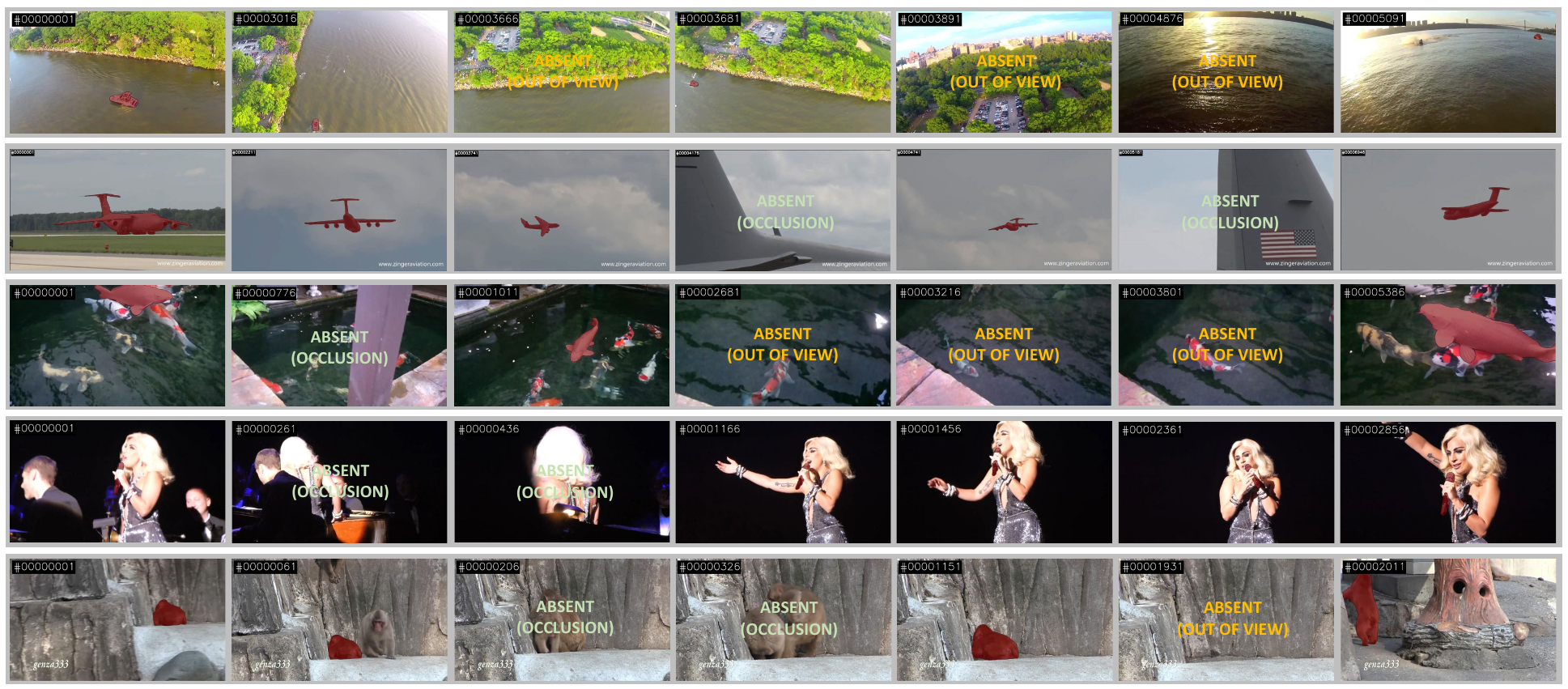}
    \vspace{-3mm}
    \caption{Example sequences of our Large-scale Long-term Video Object Segmentation (LVOS). Compared to previous video object segmentation datasets, LVOS presents greater challenges, with the main difficulties stemming from longer video durations, intricate scenes, frequent disappearance and reappearance of objects, Cross-temporal confusion, and small objects. Text in \textcolor[rgb]{0.77,0.88,0.71}{green} denotes that target object is occluded, while text in \textcolor[rgb]{1.0,0.75,0}{orange} denotes that target object is out-of-view. }
    \label{fig:data_overview}
    \vspace{-5mm}
\end{figure*}

\section{Introduction}
\label{sec:intro}

Video Object Segmentation (VOS) is a critical task in video analysis, aiming to accentuate a specific target in a given video, beginning with a designated object appearance in the initial frame. VOS is pivotal for enhancing video comprehension and holds immense potential across various domains, including video editing~\cite{oh2018fast}, augmented reality~\cite{ngan2011video}, robotics~\cite{cohen1999detecting,erdelyi2014adaptive}, self-driving cars~\cite{zhang2016instance,ros2015vision,saleh2016kangaroo}. Based on different ways of describing the target object, VOS can be divided into several settings, such as semi-supervised VOS~\cite{xu2018youtube,perazzi2016benchmark}, where the target object's mask is provided for the first frame, interactive VOS~\cite{chen2018blazingly,oh2019fast}, wherein users engage in interactions using dots or scribbles, and unsupervised VOS~\cite{perazzi2016benchmark} where no interaction is provided and models need to automatically detect salient objects. In real-world applications, VOS tasks encounter multifaceted challenges, with objects undergoing frequent disappearance and reappearance, and video duration commonly exceeding one minute. Given these complexities, it becomes imperative for VOS models to perform with precision in redetecting and segmenting target in videos of arbitrary lengths.

However, existing Video Object Segmentation (VOS) models are predominantly tailored for short-term scenarios, facing challenges when confronted with the complexities of long-term videos. These models exhibit vulnerabilities in addressing prolonged object disappearances and suffer from error accumulation over time~\cite{voigtlaender2019feelvos,yang2020collaborative,yang2021associating}. Notably, models such as~\cite{oh2019video,cheng2021rethinking,hu2021learning,xie2021efficient,seong2020kernelized} may encounter efficiency issues and out-of-memory crashes as a consequence of an ever-expanding memory bank, particularly evident in the context of lengthy videos. However, the absence of densely annotated long-term VOS datasets poses a significant constraint on the practical advancement of VOS. The performance of these models in real-world video scenes remains largely unexplored.
    
To date, nearly all VOS benchmark datasets, including DAVIS~\cite{perazzi2016benchmark} and YouTube-VOS~\cite{xu2018youtube}, just focus on short-term videos, a limitation that inadequately addresses the demands of practitioners. These datasets typically feature an average video length of less than 6 seconds, with consistently visible target objects. This characteristic starkly differs from real-world scenarios where the average duration is significantly longer (i.e., 1-2 minutes), and target objects frequently undergo disappearance and reappearance. The existing benchmarks, therefore, fail to represent the complexities inherent in practical situations, necessitating the development of long-term VOS datasets that align more closely with the challenges faced by real-world applications.

To this end, we propose the first large-scale \textbf{long-term} video object segmentation benchmark dataset, named \textbf{L}ong-term \textbf{V}ideo \textbf{O}bject \textbf{S}egmentation (\textbf{LVOS}). LVOS contains 720 videos with an average duration of 1.14 minutes. The emphasized properties of LVOS are summarised as follows. 

\begin{enumerate}
    \item{\textbf{Long-term}. Videos in LVOS persist for 1.14 minutes on average as opposed to the 6 seconds typical of short-term videos, making them more in line with real-world applications (refer to Table~\ref{tab:dataset_static} for statistic details). Fig.~\ref{fig:data_overview} shows some representative examples from LVOS. These videos cover multiple challenges, especially attributes specific to long-term videos such as frequent reappearance and sustained confusion amongst similar objects.}
    \item{\textbf{Large-Scale}. The LVOS dataset encompasses approximately 296K frames across 720 videos, exceeding the previous largest VOS dataset by twice the size of frame numbers. Each video within LVOS presents complicated challenges that necessitate a more robust and capable VOS model to effectively tackle and decipher them.}
    \item{\textbf{Dense and high-quality annotations}. Every frame within the LVOS is manually and accurately annotated at a rate of 6 FPS. To ensure precise and efficient target object annotation, we have developed a semi-automatic annotation pipeline. There are 407K annotated objects contained, more than double the count of labeled objects in YouTube-VOS~\cite{xu2018youtube} and roughly on par with that of MOSE~\cite{MOSE}. Notably, despite having approximately twice as many frames as MOSE, LVOS has essentially the same amount of annotations as MOSE, which is a good indication that there are frequent long-duration disappearances of objects in LVOS.}
    \item{\textbf{Comprehensive labeling}. The videos incorporated into LVOS span 44 categories, effectively encapsulating typical everyday scenarios. Of these 44 categories, 12 remain unseen, specifically designed to evaluate and better gauge the generalization capabilities of VOS models.}

\end{enumerate}

\begin{table*}
    \caption{Comparison of LVOS with the most popular video segmentation and tracking benchmarks. The top part is existing short-term video datasets and the bottom part is long-term video datasets. Duration denotes the total duration (in minutes) of the annotated videos. Frame rate denotes the sampling rate, measured in frames per second (FPS). For BURST, the training set is annotated at a rate of 6 FPS, while the test set and validation set are annotated at a rate of 1 FPS. Annotations type means the type of groundtruth annotations. \textbf{M} and \textbf{B} denote mask and box annotations. \textbf{N} means that the groundtruth annotations are unavailable. . The largest value is in bold, and the second and third largest values are underlined.}
    \setlength{\tabcolsep}{2.0mm} 
    \input{./tables/static}
	\centering
	\label{tab:dataset_static}
\end{table*}	

	We conduct comprehensive experiments on the LVOS dataset to evaluate the performance of existing VOS models. Specifically, we assess 20 VOS models under 4 different settings, including semi-supervised video object segmentation, unsupervised video single object segmentation, unsupervised video multiple object segmentation, and interactive video object segmentation. Despite the commendable performance of these models on short-term videos (up to about 90 $\% \mathcal{J} \& \mathcal{F}$ on YouTube-VOS~\cite{xu2018youtube}), these models suffer from a notable performance decline in long-term videos. Through attribute-based analysis and  visualization of prediction results, the poor accuracy results from complex motion, large scale variations, frequent disappearances, and similar background confusion. Then, to explore the key to improvement of accuracy on long-term video tasks, we retrain these models on LVOS training sets, and find that the diverse scenes in LVOS can obviously enhance the performance on long-term videos. Additionally, through oracle experiments, we identify that error accumulation over time is another contributing factor of the reduced performance. Through our extensive experiments and analysis, we uncover the root causes of the unsatisfied performance of these models on long-term video tasks and illuminate potential avenues for future improvement.

Our contributions can be encapsulated as follows:
\begin{enumerate}
    \item{We construct  a novel, densely and high-quality annotated, long-term video object segmentation segmentation dataset, called LVOS. LVOS encompasses 720 videos with an average duration of 1.14 minutes,  providing a comprehensive array of labels.}
    \item{We perform a series of experiments to evaluate 20 existing VOS models on LVOS under 4 different settings to provide a comprehensive evaluation of their performance. The result of these experiments indicates that the increase in video length is the primary contributing factor to performance degradation, underscoring the critical necessity and significance of our LVOS.}
    \item{We conduct a thorough analysis on the performance of current VOS models to identify and explain the reason of the poor performance of these models when dealing with long-term video tasks. We explore possible avenues to improve the accuracy of VOS models on long-term video tasks, providing insight into feasible directions for future research of VOS in the real-world scenarios.}
\end{enumerate}

It is noteworthy that this work is the extension of our previous conference version in~\cite{hong2023lvos}. The primary new contributions are enumerated below. (1) We augment the original dataset and increase the video count from 220 to 720, with more than about 170K frames and 251K annotations. This broadened video content covers a more diverse range of scenes, augmenting training efficacy and testing generalization. We also incorporate more static comparison with existing related datasets to emphasize the uniqueness and necessity of LVOS. (2) We refine the semi-automatic annotation pipeline to speed up the labeling process and provide more details about LVOS. (3) We conduct more comprehensive experiments on LVOS to thoroughly assess existing video object segmentation models under 4 different settings. We directly evaluate those VOS models, pre and post LVOS training, to demonstrate the unsatisfied performance on long-term videos. (4) We add additional analysis and visualization to explore the reason behind the inferior performance of these models on long-term videos, and discuss the potential avenues for enhancing the performance. We also present a discussion on the potential future directions and work in the domain of video object segmentation.

%% file: tables/static.tex
\begin{tabular}{l|cccccccccc}
			\toprule
			Dataset   & Videos & \makecell[c]{Mean \\ Frames} & \makecell[c]{Total \\ Frames} & \makecell[c]{Mean \\ Duration}  & \makecell[c]{Total \\ Duration} & \makecell[c]{Frame \\ Rate } & \makecell[c]{Object \\ Classes} & Objects & Annotations & \makecell[c]{Annotations \\ Type}    \\
			\midrule
            \multicolumn{11}{c}{Short-term Video Datasets} \\
            \midrule
            
			FBMS~\cite{ochs2013segmentation}       & 59            & 235          & 13,860         & 0.13          & 7.7           & 30               & 16              & 139     & 1,465      & \textbf{M}             \\
			DAVIS~\cite{pont20172017}       & 90          & 69          & 6,298         & 0.04          & 5.17           & 24               & -              & 205     & 13,543      & \textbf{M}      \\
			YouTube-VOS~\cite{xu2018youtube} & \textbf{4,453}       & 27          & 120,532       & 0.06          & 334.8         & 6                & \underline{94}             & \underline{7,755}   & 197,272     & \textbf{M}         \\
			YouTube-VIS~\cite{yang2019video} & \underline{2,883}          & 28          & 78,000      & 0.06          & 216.7         & 6                & 40          & 4,883  & $ \sim$131,000    & \textbf{M}            \\	
			OVIS~\cite{qi2022occluded}  & 901          & 90         & $ \sim$68,650      & 0.21       &190.7         & 6                & 25          & 5,223  &  $\sim$296,000     & \textbf{M}            \\	
			UVO~\cite{wang2021unidentified}  & 1,200          & 28          & $ \sim$108,000    & 0.05          & 511        & 30               & -         & \textbf{14,748}  &  \underline{$\sim$1,327,000}   & \textbf{M}            \\	
			VOT-ST 2021~\cite{kristan2021ninth} & 60       & 324        & 19,447     & 0.18      & 10.8        & 30    & -         & 60  & 19,379   & \textbf{M}       \\
			VOT-ST 2022~\cite{kristan2022tenth} & 62       & 321        & 19,903     & 0.18      & 11.1        & 30    & -         & 62  & 19,826   & \textbf{M}       \\          
            BURST~\cite{athar2023burst} & \underline{2,914}      & 214        & \underline{624,240}     & 0.60          & \underline{1,734}        & 6 / 1     & \textbf{482}         & \underline{16,089}  &  \underline{600,157}   & \textbf{M}          \\

            MOSE~\cite{MOSE} & 2,149       & 73        & $\sim$159,600     & 0.21          & 443.6        & 6               & 36         & 5,200  & 431,725   & \textbf{M}          \\	
			\midrule
            \multicolumn{11}{c}{Long-term Video Datasets} \\
            \midrule
			VOT-LT 2019~\cite{kristan2019seventh}  & 	50       & \textbf{4,305}    & 215,298         & \textbf{2.39}           & 119              & 30               & -              & 50       & 215,298            & \textbf{B} \\
            VOT-LT 2022~\cite{kristan2022tenth}  & 	50       & \underline{3,366}    & 168,282         & \underline{1.87}          & 93              & 30               & -              & 50       & 168,282            & \textbf{B}                \\
            VOT-ST 2023~\cite{Kristan_2023_ICCV} & 144       & 2,073        & 298,640     & 1.15      & 166        & 30    & -         & 341  & -   & \textbf{N}       \\	

			UAV20L~\cite{mueller2016benchmark}     & 	20       & \underline{2,934}   & $\sim$59,000         & \underline{1.63}          & 32.6            & 30               & 5             & 20       & $\sim$59,000             & \textbf{B}               \\
			LaSOT~\cite{fan2019lasot}  & 	1,500         & 2,502    & \textbf{$\sim$3,870,000}       & 1.39           & \textbf{2,148}             & 30               & \underline{85}              & 1,550      & \textbf{$\sim$3,870,000}           & \textbf{B}               \\
			YouTube-VIS 2022 Long~\cite{vis2022}  & 121     & 75         & 9,014      & 0.8         & 100         & 1.5                &-         & -  &  -    & \textbf{N}          \\	
			YouTube-VOS 2022 Long~\cite{vos2022} & 116        & 67         &7,873     & 0.74          & 87        & 1.5               &-         & 116  &  -    & \textbf{N}           \\	
			Long-time Video~\cite{liang2020video}  & 3           & 2,470        & 7,411         & 1.3           & 4              & 30               & -              & 3       & 60          & \textbf{M} \\
            \midrule
			LVOS V1~\cite{hong2023lvos}        & 220         & 574         & 126,280       & 1.59           & 351           & 6                & 27             & 282     & 156,432      & \textbf{M}   \\
            \textbf{LVOS V2}        & 720         & 412         & \underline{296,401}       & 1.14           & \underline{823}            & 6                & 44             & 1,132     & 407,945      & \textbf{M}   \\
			\bottomrule         
		\end{tabular} 

%% file: section/2_related.tex
\section{Related work}
\subsection{Short-term Video Object Segmentation Dataset.} All existing benchmark datasets for video object segmentation (VOS) are short-term video datasets.
FBMS~\cite{ochs2013segmentation} comprises 59 sequences with a total of 13,860 frames, divided into 29 and 30 videos for the training and evaluation sets, respectively. DAVIS 2017~\cite{pont20172017}, a widely recognized benchmark dataset, includes 60 and 30 videos for the training and validation sets, respectively, with a total of 6,298 frames. Each frame in DAVIS 2017 is annotated with pixel-level precision and high quality. YouTube-VOS~\cite{xu2018youtube}, as a large-scale dataset, contains 3,252 sequences with precise annotations at 6 FPS and covers 78 diverse categories.  MOSE~\cite{MOSE} a recently released dataset for complex video object segmentation scenarios, consists of 2,149 videos. Videos in MOSE are much more challenging than those in DAVIS and YouTube-VOS, with objects frequently disappearing or being occluded. All these benchmarks are short-term video datasets, with the average video duration ranging from 3 to 10 seconds.
Despite some VOS methods~\cite{liang2020video,li2020fast,li2022recurrent,cheng2022xmem} claiming to scale well to long-term videos, they have not conducted quantitative experiments on a long-term VOS benchmark due to the absence of such a dataset. Videos in LVOS are long-term, with an average duration of approximately 1.14 minutes, making them more applicable to real-world scenarios.

\subsection{Long-term Tracking Dataset.}
Several benchmark datasets are specifically designed for long-term tracking. UAV20L~\cite{mueller2016benchmark} is a small-scale dataset comprising only 20 long videos. OxUvA~\cite{valmadre2018long} consists of 366 sequences, however, each video is sparsely annotated at a frame rate of 30 FPS. LaSOT~\cite{fan2019lasot}, the first large-scale and densely annotated long-term tracking dataset, provides 1,400 videos totaling 3.52M frames. The sequences in LaSOT average 2,512 frames at 30 FPS, with each frame manually annotated with a bounding box. These long-term tracking datasets underscore the importance of long-term tasks. Nonetheless, these datasets only provide box-level annotations, and pixel-level annotations are unavailable, which is more crucial for fine-grained study. Long-time Video~\cite{liang2020video} comprises three extended videos, averaging 2,470 frames per video, with only 20 frames uniformly annotated for each video.
It's worth noting that YouTube-VOS 2022 Long~\cite{vos2022} and YouTube-VIS 2022 Long~\cite{vis2022}, proposed at the CVPR 2022 workshop, also include long-term videos, while no groundtruth is available. A comprehensive long-term VOS dataset, complete with training data and constant availability, is lacking. Thus, we provided LVOS, which focuses on long-term video object segmentation. LVOS comprises a total of 220 videos, including training, validation, and test sets. Each frame in LVOS is manually and precisely annotated. We propose LVOS to foster the development of robust VOS models and provide a more suitable evaluation benchmark for practical application.

\begin{figure*}[htbp]
		\centering
		\includegraphics[width=1.0\linewidth]{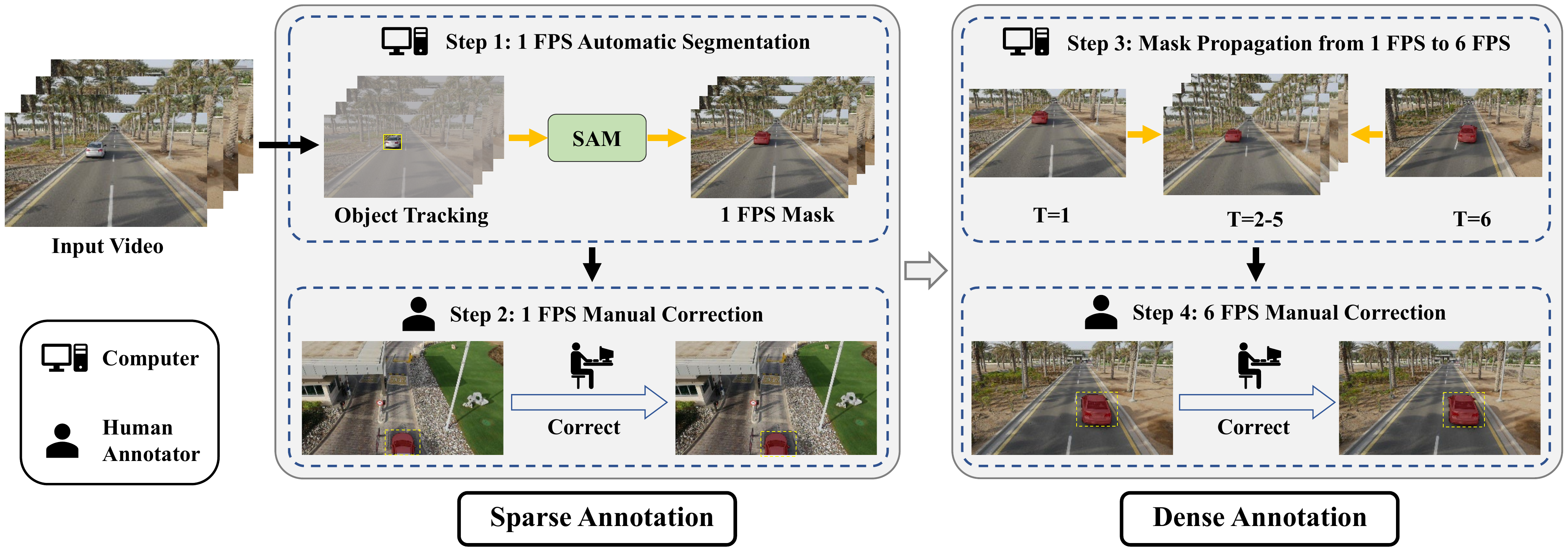}
		\caption{Annotation Pipeline, including four steps. Step 1: 1 FPS Automatic Segmentation. Object tracking~\cite{wei2023autoregressive} models and SAM~\cite{kirillov2023segment} are adopted to automatically segment the target object at 1 FPS. Step 2: 1 FPS Manual Correction. We refine and correct masks obtained in Step 1 manually. Step 3: Mask Propagation from 1 FPS to 6 FPS. We propagate masks from 1 FPS to 6 FPS by utilizing a VOS model~\cite{yang2022decoupling}. Step 4: 6 FPS Manual Correction. We manually correct the masks obtained in Step 3.}
		\label{fig:anno_pipeline}
\end{figure*}

\subsection{Video Object Segmentation.} 
Based on the manner employed to propose the target object, video object segmentation can be divided into four categories.

\textbf{Unsupervised Video Object Segmentation}.
Unsupervised Video Object Segmentation (UVOS), which aims to identify noteworthy objects within a video, does not require any annotations. Current UVOS models commonly detect the target object by leveraging the motion cues between adjacent frames. ~\cite{wang2019learning,lu2020video,lu2019see,wang2019zero,lu2021segmenting} harness the high consistency of visual attention from preceding images to distinguish salient objects directly. ~\cite{zhou2020motion,ren2021reciprocal,ji2021full,yang2021learning,zhang2021deep,pei2022hierarchical,cho2023treating,hong2023simulflow} take advantage of the optical flow for target recognition. These models typically conduct evaluation on short-term video datasets, such as DAVIS, FBMS, and YouTube-Objects~\cite{prest2012learning}. 
While these models leverage local temporal information and deliver remarkable precision on short-term videos, their performance on long-term videos is far from satisfactory due to insufficient incorporation of global temporal context.

\textbf{Semi-supervised Video Object Segmentation}.
Semi-supervised Video Object Segmentation (VOS) identifies, tracks, and segments the target object through the entire video based on its appearance provided in the first frame. The key to semi-supervised VOS is constructing and utilizing a feature memory.~\cite{caelles2017one,maninis2018video,voigtlaender2017online,xiao2018monet,robinson2020learning, meinhardt2020make, park2021learning, bhat2020learning} implement online learning methodologies to finetune pretrained networks at test time, requiring a large amount of time.~\cite{hu2018videomatch,chen2018blazingly,shin2017pixel,bao2018cnn} utilize the manually annotated first frame to guide the segmentation of subsequent frames. Conversely, ~\cite{perazzi2017learning,zhang2019fast,cheng2018fast,jang2017online,chen2020state,oh2018fast,xu2019spatiotemporal,hu2018motion,ventura2019rvos,hu2017maskrnn,li2018video,wang2019fast,khoreva2017lucid} employ the previously segmented frame as a reference to facilitate a frame-to-frame mask propagation.~\cite{voigtlaender2019feelvos,yang2020collaborative,yang2018efficient,wang2019ranet,johnander2019generative,hong2021adaptive} merge both the first and preceding frames to serve as a feature memory, yet the temporal context thus provided remains limited. To overcome this constraint,~\cite{oh2019video,9665289,cheng2021modular,seong2020kernelized,hu2021learning,xie2021efficient,wang2021swiftnet,lu2020video,cheng2021rethinking,seong2021hierarchical,mao2021joint,yang2021associating,park2022per,lin2022swem,guo2022adaptive} develop a feature memory bank to save all previous frames. However, this perpetually expanding memory bank may confront an out-of-memory crash when processing long-term videos.
	
Recently, several studies have begun to focus on the specific challenges posed by long-term videos.~\cite{liang2020video} crafts an adaptive feature bank, utilizing exponential moving averages for dynamic management of crucial object features. Global context module is proposed by~\cite{li2020fast} to effectively summarize target information.~\cite{li2022recurrent} builds the memory bank of fixed size by utilizing a recurrent dynamic embedding (RDE). Xmem~\cite{cheng2022xmem} develops three types of memory banks interconnected to segment the current frame. Through compressing the feature bank, these methods succeed in maintaining constant memory cost, however, they still struggle with tracking loss after a lengthy period of object disappearance in long-term videos. While these studies attempt to address challenges in long-time video object segmentation, the results remain unsatisfactory.

\textbf{Referring Video Object Segmentation}.
Referring video object segmentation (RVOS) is a multimodal task designed to distinguish and segment the corresponding object in each frame depending on a given language description. This process necessitates the use of appearance and motion information. 3D ConvNets (\textit{e.g.}, I3D~\cite{carreira2017quo}) are utilized to model the temporal feature and mine the semantic correspondence between visual and linguistic features in~\cite{gavrilyuk2018actor,mcintosh2020visual,ning2020polar,wang2020context,wang2019asymmetric,ye2021referring}. URVOS~\cite{seo2020urvos} employs cross-model attention to leverage multimodal information. Memory mechanism is taken advantage of to combine referring object segmentation and mask propagation and improves performance greatly. In~\cite{zhao2022modeling,ding2022language,yang2022actor}, optical flow or frame differences are utilized to extract the motion feature between adjacent frames. MTTR and ReferFormer~\cite{wu2022language,botach2022end} adopt transformers to perform RVOS, where the attention mechanism is used to combine the linguistic and visual features and model the temporal relation.

\textbf{Interactive Video Object Segmentation}.
In the previous three types of video object segmentation tasks, human interaction with the model is limited to a maximum of one time, which means segmentation errors can't be rectified. Interactive Video Object Segmentation (IVOS) is proposed to implement segmentation and error correction of target objects through multiple human interactions, such as clicks or scribbles. Existing IVOS models~\cite{cheng2021modular,oh2019fast,miao2020memory,yin2021learning,chen2018blazingly,cheng2018fast,heo2021guided}, which are built upon semi-supervised video object segmentation methods, start by converting the initial interaction into a per-frame segmentation. Then, interactive modules are designed to correct the prediction error in a specific frame based on the user's interaction, and mask propagation modules propagate the corrected information to other frames. Unlike the previous three kinds of video object segmentation tasks, IVOS models achieve enhanced performance due to increased interaction. However, IVOS models perform also poorly when dealing with long-term videos. Compared to short videos, direct propagation of a corrected mask to all frames in a long video can worsen segmentation results due to the complex motion in longer videos. Besides, current IVOS models require the storage of all masks and visual features for each frame, which may lead to memory overflow. Clipping video into several shots may result in the loss of important temporal information. Therefore, the development and evaluation of IVOS algorithms capable of handling long-duration videos is of significant importance.

%% file: section/3_method.tex
\section{LVOS: Large-scale Long-term Video Object Segmentation Benchmark Dataset}
Our principal objective in developing the LVOS is to establish an expansive benchmark explicitly designed for the needs of long-term video object segmentation, which reflects realistic applications more accurately with multiple challenging attributes. In Section~\ref{sec:data_construct}, we introduce the construction of LVOS. Section~\ref{sec:anno_pipeline} encompasses an overview of the semi-automated annotation process we have developed. Later, in Section~\ref{sec:dataset_static}, we present a comprehensive dataset statistics and attributes.

\begin{figure}[htbp]
	\centering
	\includegraphics[width=1.0\linewidth]{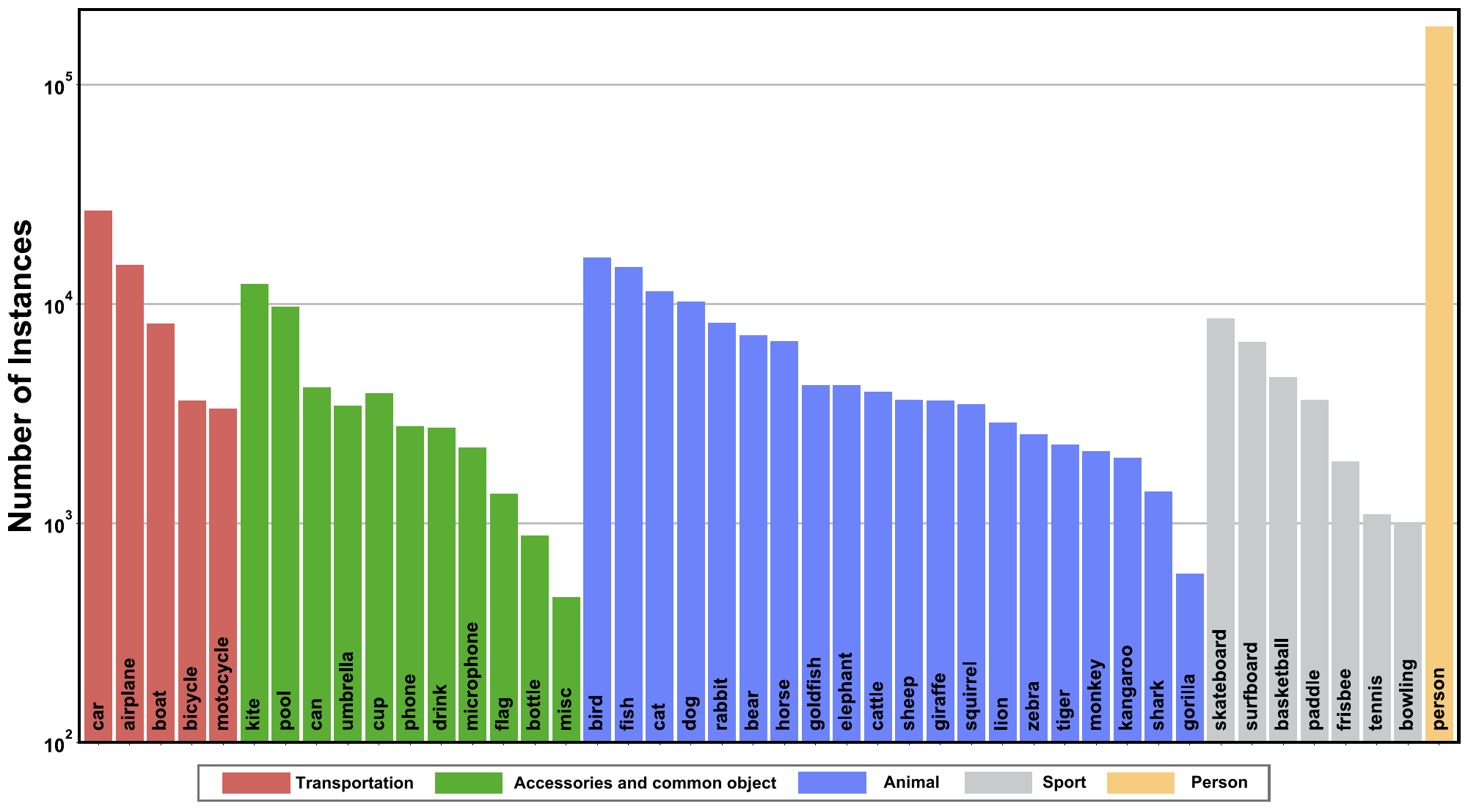}
	\caption{The histogram of instance masks for five parent classes and sub-classes. Objects are sorted by frequency. The entire category set roughly covers diverse objects and motions that occur in everyday scenarios. }
	\label{fig:instance_nums}
\end{figure}

\subsection{Dataset Construction}
\label{sec:data_construct}
	\textbf{Dataset Design.} To address the shortage of dedicated resources, LVOS aims to offer the research community a novel VOS dataset for training and evaluating robust VOS models. We adhere to the following four guiding principles in the course of LVOS's construction.

\begin{enumerate}

    \item{\textbf{Long-term}. In comparison to existing VOS datasets~\cite{perazzi2016benchmark,xu2018youtube,MOSE} where the average video duration is merely 3-10 seconds, LVOS ensures a considerably longer duration. Specifically, videos within LVOS last approximately 1.14 minutes (\textit{i.e.}, 412 frames at 6 FPS), constituting about six times the duration of short-term videos. Consequently, LVOS provides a more realistic representation of application scenarios. Fig.~\ref{fig:data_overview} displays some representative examples, including multiple challenges, particularly those unique to long-term videos.}
    \item{\textbf{Large-Scale}. With a total collection of 297K frames across 720 videos, LVOS significantly surpasses current VOS datasets, exceeding them by at least a factor of two. Each video within LVOS represents complex challenges. We believe that a large-scale dataset plays an important role in training and evaluating video object segmentation models in real scenes.}
    \item{\textbf{Dense and high-quality annotations}. Due to the time-consuming mask annotation, the duration and scale of current VOS datasets are constrained to a great extent. Thus, we develop a semi-automatic annotation pipeline to annotate each frame efficiently and concisely at a frame rete of 6 FPS. High-quality and densely annotated masks are pivotal to the training of VOS models and for assessing their performance in real-world applications.}
    \item{\textbf{Comprehensive labeling}. We design a set of categories that are pertinent to daily life with 5 parent classes and 44 subclasses. Notably, the 44 categories are not limited to COCO dataset~\cite{lin2014microsoft}, featuring some categories not present in the COCO dataset, such as 'frisbee'. Among the 44 categories, there are 12 unseen categories to better evaluate the generalization ability of models.}

\end{enumerate}

	\textbf{Data Collection.}
	To construct LVOS, we carefully select a set of categories comprising 5 parent classes and 44 subclasses from the videos in existing datasets, including VOT-LT 2019~\cite{kristan2019seventh}, VOT-LT 2022~\cite{kristan2022tenth}, VOT-ST 2023~\cite{Kristan_2023_ICCV}, LaSOT~\cite{fan2019lasot}, AVisT~\cite{noman2022avist}, ARKitTrack~\cite{zhao2023arkittrack}, UVOT400~\cite{alawode2023improving}, DanceTrack~\cite{sun2022dancetrack}, BURST~\cite{athar2023burst}, and OVIS~\cite{qi2022occluded}. These datasets contain more than 6,500 videos in total. VOT-LT 2019~\cite{kristan2019seventh}, VOT-LT 2022~\cite{kristan2022tenth}, and LaSOT~\cite{fan2019lasot} have been customized for long-term tracking. OVIS~\cite{qi2022occluded} is designed for video instance segmentation in complex scenarios. BURST~\cite{athar2023burst} and VOT-ST 2023~\cite{Kristan_2023_ICCV} are proposed for video object segmentation and tracking, but the videos in these two datasets either lack mask annotations or have only sparse mask annotations at 1 FPS or the duration of videos is too short. AVisT~\cite{noman2022avist}, ARKitTrack~\cite{zhao2023arkittrack}, UVOT400~\cite{alawode2023improving}, DanceTrack~\cite{sun2022dancetrack} are proposed for tracking tasks in challenging scenes, but only box annotations are available. Thus, we screen about 1,500 videos with a resolution of 720P as candidate videos, ensuring that the videos last for more than 1 minutes and pose complex challenges. Finally, after comprehensive consideration of video quality and tracking difficulty, 720 videos are selected to constitute LVOS. For target selection, we may either follow the target object in the original datasets, or select different objects as targets.

\begin{table}[htbp]
	\centering
	\caption{Definitions of video attributes in LVOS. We extend and modify the short-term video challenges defined in \cite{perazzi2016benchmark} (top), which is exanded with a complementary set of  long-term video attributes (bottom).}
	\label{tab:challenge}
	\input{./tables/attribute}
\end{table}

\subsection{Semi-Automatic Annotation Pipeline}
	\label{sec:anno_pipeline}
    The labor-intensive mask annotation process poses a significant constraint on the scale of VOS dataset. To address this, we develop a semi-automatic annotation pipeline to streamline the frame annotation process. This pipeline encompasses four steps, which are illustrated in Fig.~\ref{fig:anno_pipeline}.
	
	\textbf{Step 1: 1 FPS Automatic Segmentation.} 
	Firstly. we manually identify the bounding box of the target objects in the first frame and then apply ARTrack~\cite{wei2023autoregressive} to propagate the box to all subsequent frames. Then, we segment the target object in each frame at 1 FPS, based on the predicted bounding boxes by using SAM~\cite{kirillov2023segment}. 
	
	\textbf{Step 2: 1 FPS Manual Correction.} The potential for tracking errors, segmentation defects, and other prediction mistakes may result in inaccuracies or the omission of the target object mask in some frames. Hence, we employ EISeg~\cite{hao2021edgeflow} (an Efficient Interactive Segmentation Tool based on PaddlePaddle~\cite{ma2019paddlepaddle}) to correct masks. Approximately 20\% of frames require correction.
	
	\textbf{Step 3: Mask Propagation.} We adopt a VOS model (\textit{i}.\textit{e}., DeAOT~\cite{yang2022decoupling}) to propagate the 1 FPS annotated masks acquired in Step 2 to their neighboring unlabeled frames, thereby automatically extending the masks from 1 FPS to 6 FPS.
	
	\textbf{Step 4: 6 FPS Manual Correction.} Given the potential errors in masks generated by the VOS model, we manually rectify each frame until the results meet our standards. In this step, around 35\% of frames necessitate further refinement. 
	
	\textbf{Time and Quality Analysis.} To evaluate the quality of our annotations, we select a random sample of 100 videos from HQYouTube-VIS~\cite{ke2022mask} training set and relabel them using our semi-automatic annotation pipeline. The resulting annotations were compared with the ground truth, yielding an average IoU score of 0.95. This score attests to the high consistency between our pipeline's annotation results and the ground truth, thereby validating the effectiveness of our pipeline. Furthermore, we ask annotators to record the total time overheads. On average, it takes an annotator 60 minutes to label an entire long-term video (500 frames at 6 FPS) using our pipeline, whereas a skilled annotator would spend 1500 minutes labeling the same video (3 minutes per frame). Hence, our pipeline significantly reduces labeling costs while maintaining annotation quality.

\begin{figure}[!t]
	\centering
	\subfloat[]{
		\includegraphics[width=0.62\linewidth]{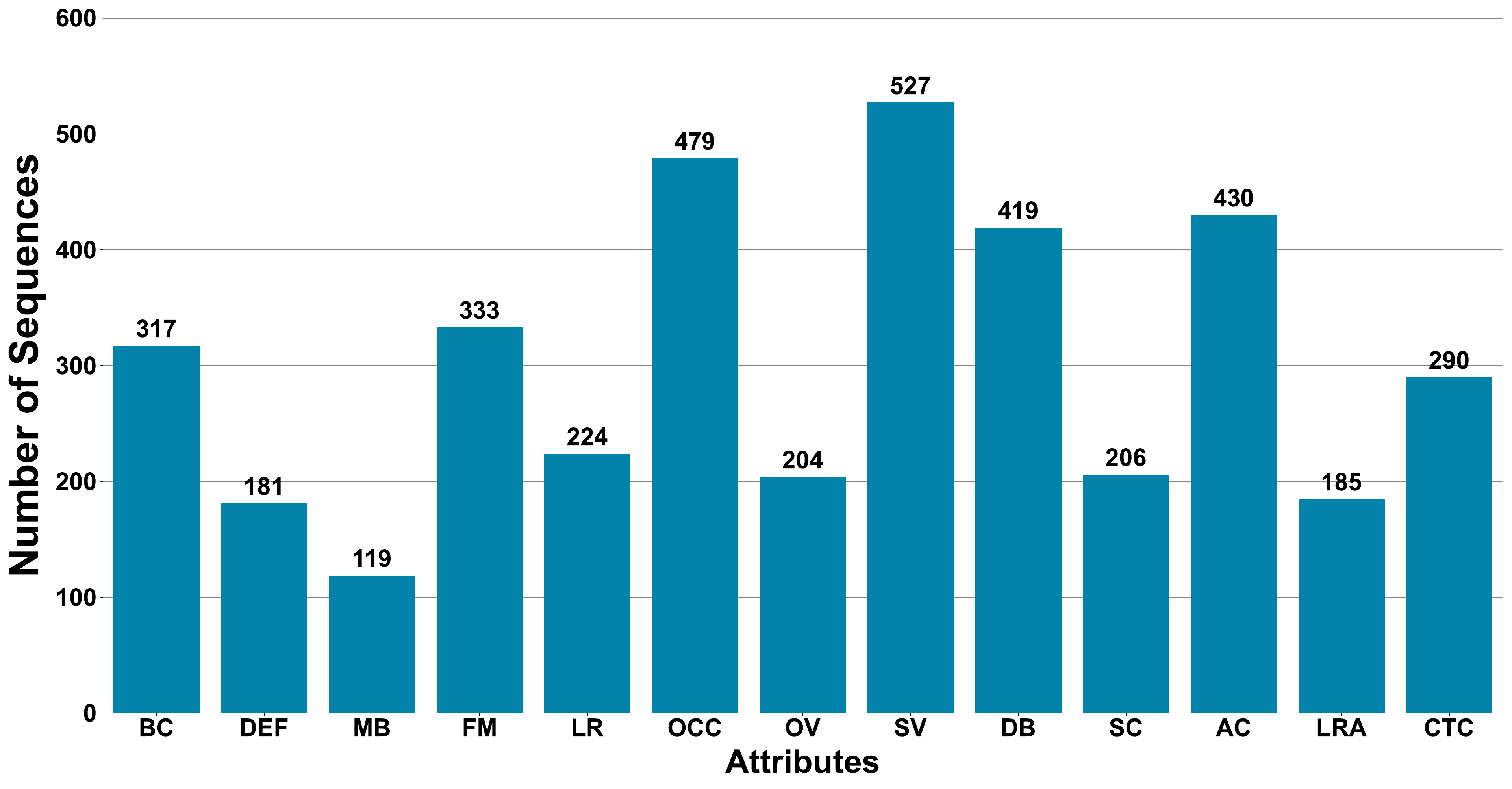}%
		\label{fig:data_attribute_table}
	}
	\hfil
	\subfloat[]{
		\includegraphics[width=0.28\linewidth]{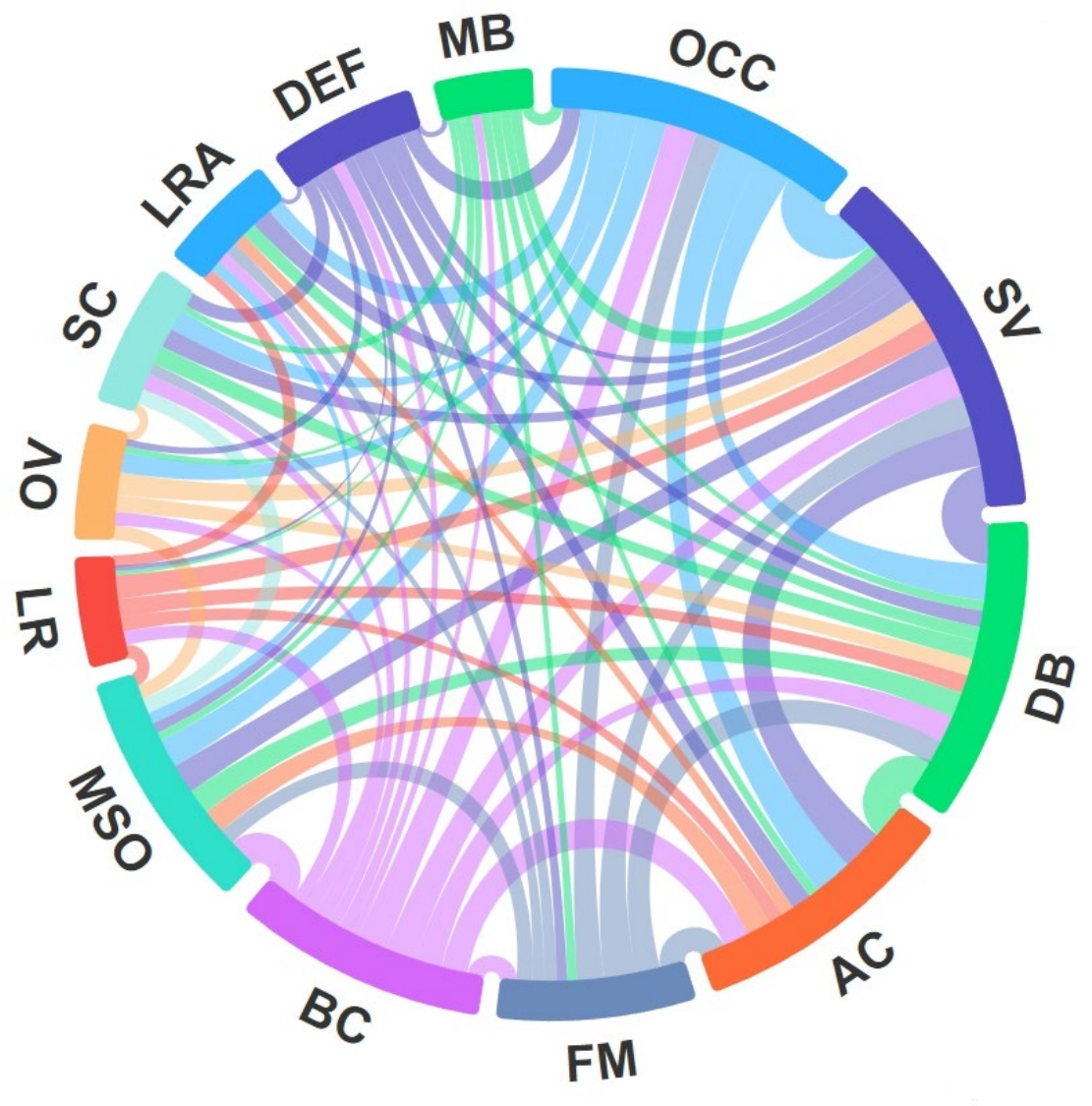}%
		\label{fig:data_attribute_chart}
	}
	\hfil
	\caption{Attributes distribution in LVOS. In sub-figure (b), the link indicates the high likelihood that more than one attributes will appear in a sequence. Best viewed in color.}
	\label{fig:data_attribute}
\end{figure}

\begin{figure*}[!t]
    \centering
    \subfloat[]{
        \includegraphics[width=0.45\linewidth]{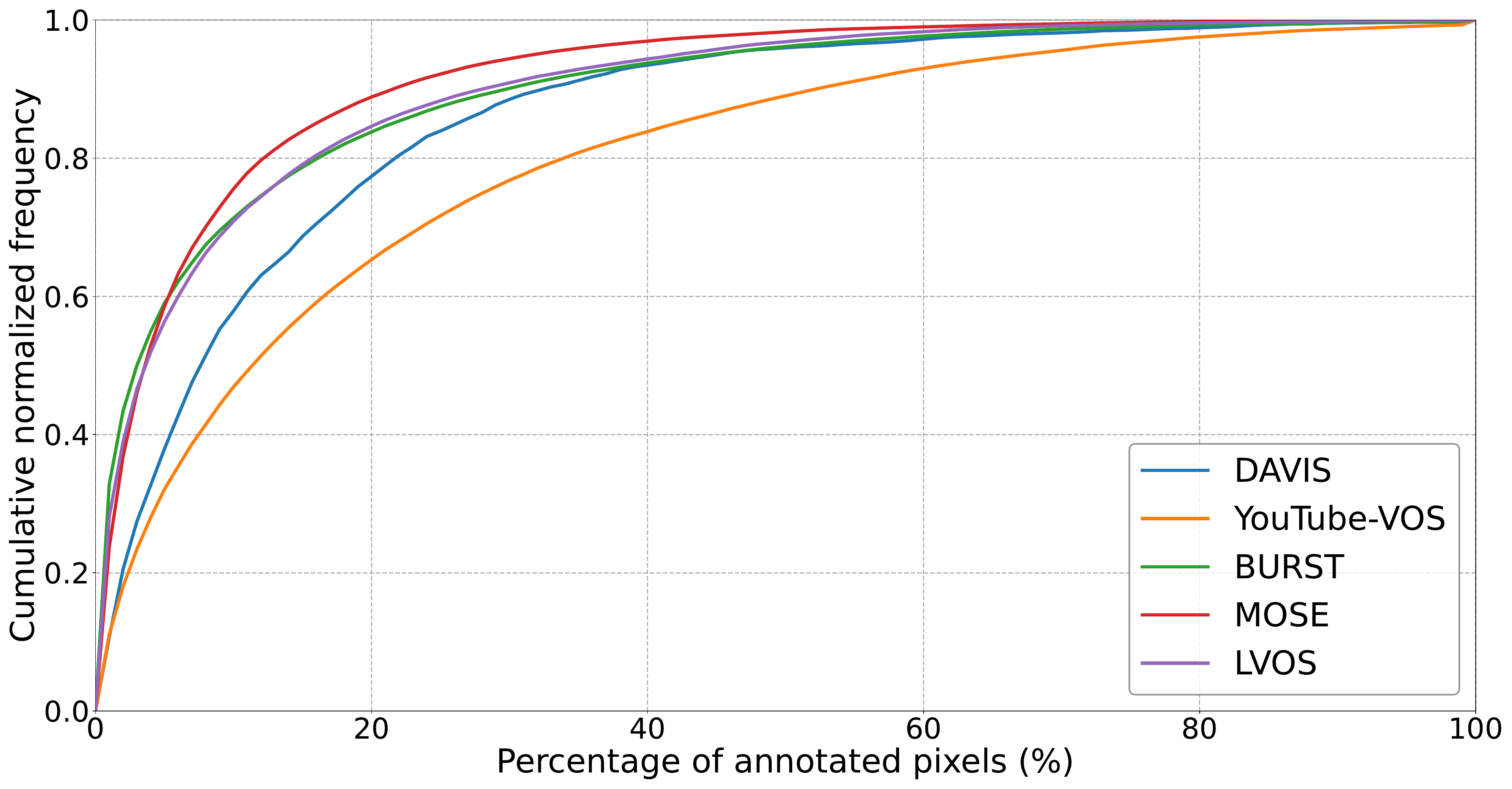}%
        \label{fig:mask_curve_all}
    }
    \hfil
    \subfloat[]{
    \includegraphics[width=0.45\linewidth]{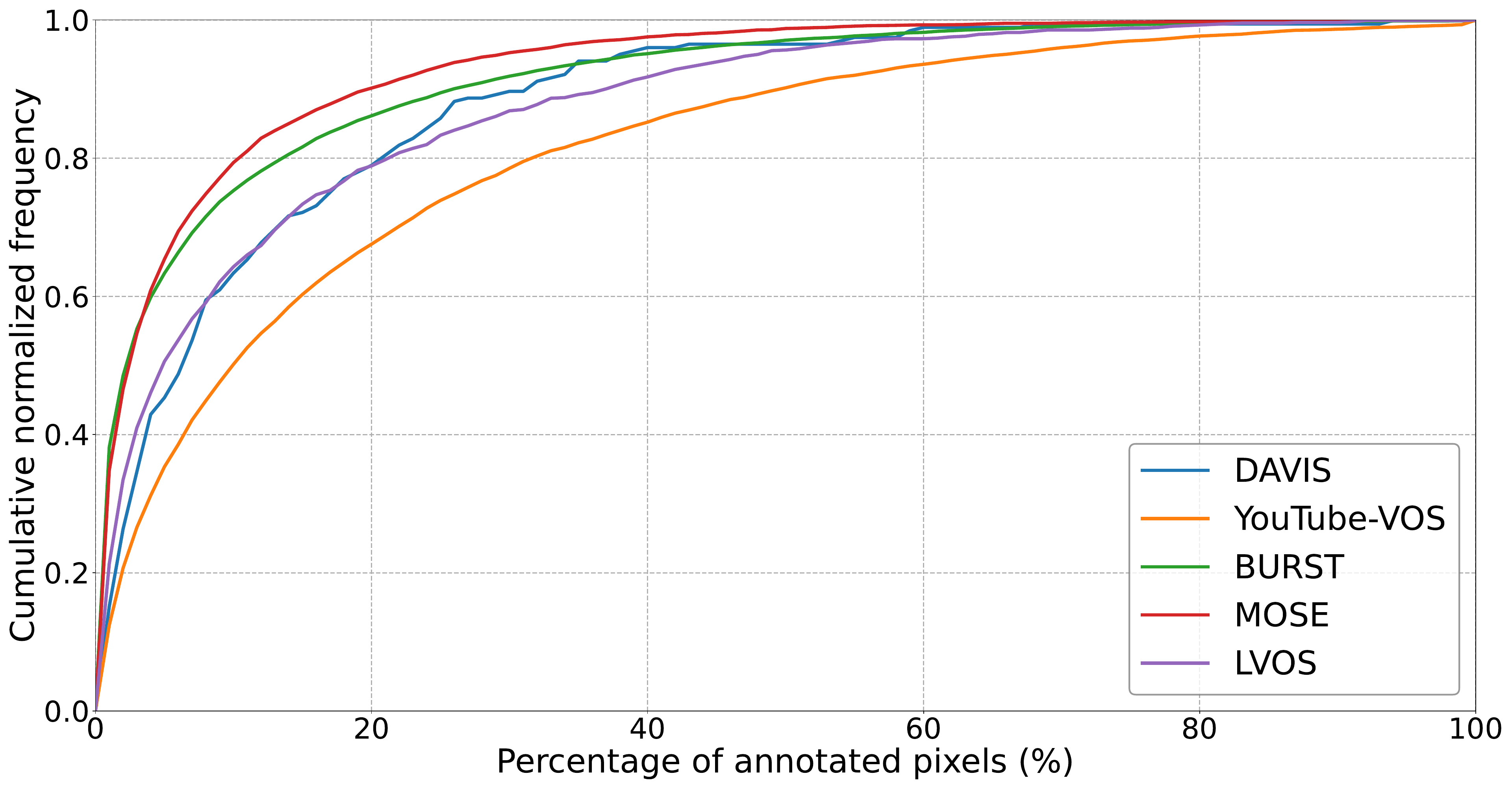}%
    \label{fig:mask_curve_first}
    }
    \hfil 
    \caption{Cumulative frequency graph of target box areas (expressed as percentages of the total image area) for different datasets. (a) displays the cumulative frequency graph based on annotations from all frames. (b) shows the cumulative frequency graph of the first frame annotations.}
    \label{fig:mask_curve_total}
\end{figure*}

\begin{figure*}[htbp]
		\centering
		\includegraphics[width=1.0\linewidth]{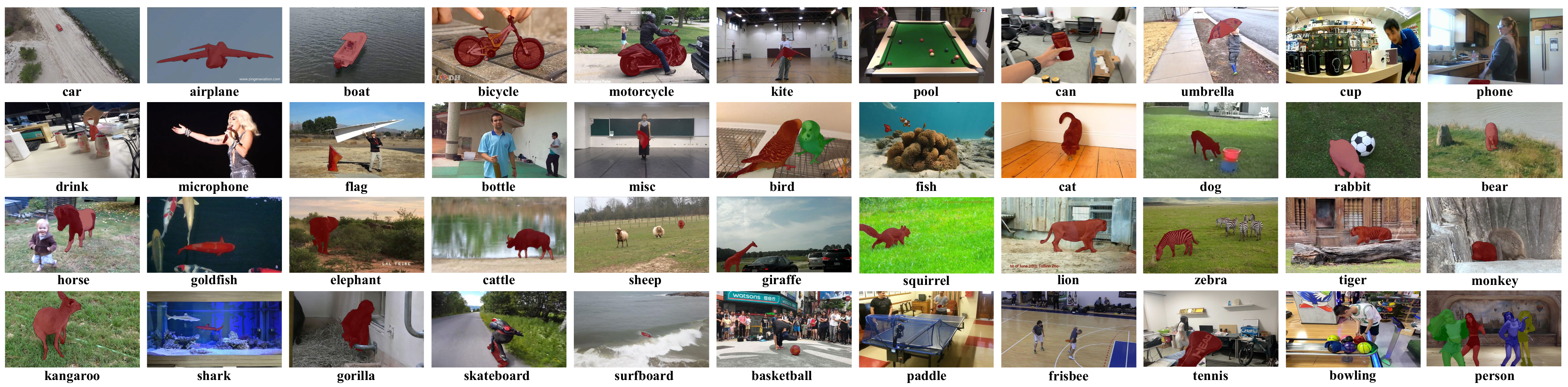}
		\caption{Example of each category. The persons or objects masked in color is the target objects in the video. Best viewed in color.}
		\label{fig:example_class}
\end{figure*}

\subsection{Dataset Statics}
\label{sec:dataset_static}
\textbf{Video-level Statics.}
	Table~\ref{tab:dataset_static} presents the video-level information of LVOS. LVOS comprises 720 videos, each with an average duration of 1.14 minutes, equating to approximately 412 frames at a frame rate of 6 FPS (\textit{vs} 3-10 seconds in short-term dataset). There are a total of 296,401 frames and 407,945 annotations, providing at least twice as many frames as other datasets~\cite{perazzi2016benchmark,xu2018youtube,liang2020video,ochs2013segmentation,MOSE}. Videos are categorized into 5 parent classes and 44 subclasses. Fig.~\ref{fig:instance_nums} provides a detailed overview of the distribution of instance masks. Notably, there are 12 categories which are not present in the training set to better evaluate the generalization ability of VOS models. We sample frames at a frame rate of 6 FPS. The videos are divided into training, validation, and testing subsets, consisting of 420, 140, and 160 videos respectively, maintaining the distribution of subsets and video length. Annotations for the training and validation sets are publicly available for the development of VOS methods, while annotations of the testing set are kept private for competition use. 	

\textbf{Attributes.}
	To undertake a further and comprehensive analysis of VOS approaches, it is critically important to identify video attributes. Accordingly, we label each sequence with 13 challenges, as defined in Table~\ref{tab:challenge}. These attributes include short-term video challenges, which are extended from DAVIS \cite{perazzi2016benchmark}, and are expanded with a complementary set of challenges specific to long-term videos. It is worth noting that these attributes are not mutually exclusive, and a single video may contain multiple challenges. Fig.~\ref{fig:data_attribute_table} and~\ref{fig:data_attribute_chart} illustrate the distribution of each video and the mutual dependencies. Scale variation (SV), occlusion (OCC), appearance change (AC), and dynamic background (DB) are the most common challenges in LVOS. Due to the extended duration of videos, both object motion and background changes are considerably more complex and varied, aspects not commonly encountered in short-term videos. The cumulative frequency graph of annotations is displayed in Fig.~\ref{fig:mask_curve_total}. The broad distribution of target sizes within our LVOS also suggests significant object size transformation in our dataset. The variation in the distribution of attributes underscores differences and higher requirements necessary for the design of VOS models.

%% file: tables/attribute.tex
\begin{tabular}{ll}
			\toprule
			Attribute & Definition  \\
			\midrule
			BC        &  \textit{Background Clutter.} The appearances of background and tar-\\  
			& get object are similar.  \\
			DEF       & \textit{Deformation.} Target appearance deform complexly.  \\
			MB        & \textit{Motion Blur.} Boundaries of target object is blurred because \\ 
			&  of camera or object fast motion.\\
			FM        & \textit{Fast Motion.} The per-frame motion of target is larger than  \\
			&   20 pixels, computed as the centroids Euclidean distance.\\
			LR        & \textit{Low Resolution.} The average ratio between target box area   \\ 
			&  and image area is smaller than 0.1 . \\
			OCC       & \textit{Occlusion.} The target is partially or fully occluded in video. \\
			OV        & \textit{Out-of-view} The target leaves the video frame completely.\\
			SV        & \textit{Scale Variation} The ratio of any pair of bounding-box is  \\
			& outside of range {[}0.5,2.0{]}.                  \\
			DB        & \textit{Dynamic Background} Background undergos deformation. \\
			SC        & \textit{Shape Complexity} Boundaries of target object is complex. \\
			AC        & \textit{Appearance Change} Significant appearance change, due to   \\      
			& rotations and illumination changes . \\
			\midrule
			LRA       & \textit{Long-term Reappearance} Target object reappears after dis- \\
			&  appearing for at least 100 frames. \\
			CTC       & \textit{Cross-temporal Confusion} There are multiple different obj-    \\
			&  ects that are similar to targect object but do not appear at   \\
            & the same time. \\
			\bottomrule            
		\end{tabular} 

%% file: section/4_experiment.tex
\begin{table*}[!t]
        \caption{Results of semi-supervised video object segmentation models on validation and test set. Subscript $s$ and $u$ denote scores in seen and unseen categories. \textcolor{blue}{$\downarrow$} represents the performance of the declining values compared to the YouTube-VOS dataset~\cite{xu2018youtube}. Mem denotes the maximum GPU memory usage (in GB). We re-time these models on our hardware (a 3090 GPU) for a fair comparison.}
        \label{tab:performance_semi_before}
        \centering
        \setlength{\tabcolsep}{2.67mm}
        \input{./tables/performance_semi_before}
    \end{table*}
    
    \begin{table*}[!t]
        \caption{Results of unsupervised video single object segmentation models on validation and test set. Subscript $s$ and $u$ denote scores in seen and unseen categories. \textcolor{blue}{$\downarrow$} represents the performance of the declining values compared to the DAVIS 2016 dataset~\cite{perazzi2016benchmark}. Mem denotes the maximum GPU memory usage (in GB). We re-time these models on our hardware (a 3090 GPU) for a fair comparison.}
        \label{tab:performance_uvos_s_before}
        \centering
        \setlength{\tabcolsep}{2.67mm}
        \input{./tables/performance_uvos_s_before}
    \end{table*}
    
    \begin{table*}[!t]
        \caption{Results of unsupervised video multiple object segmentation models on validation and test set. Subscript $s$ and $u$ denote scores in seen and unseen categories. \textcolor{blue}{$\downarrow$} represents the performance of the declining values compared to the DAVIS 2017 dataset~\cite{pont20172017}. Mem denotes the maximum GPU memory usage (in GB). We re-time these models on our hardware (a 3090 GPU) for a fair comparison.}
        \label{tab:performance_uvos_m_before}
        \centering
        \setlength{\tabcolsep}{2.67mm}
        \input{./tables/performance_uvos_m_before}
          \vspace{2mm}
    \end{table*}
    
    \begin{table*}[!t]
        \caption{Results of interactivate video object segmentation models on validation and test set. Subscript $s$ and $u$ denote scores in seen and unseen categories. \textcolor{blue}{$\downarrow$} represents the performance of the declining values compared to the DAVIS-19 dataset~\cite{caelles20192019}. Mem denotes the maximum GPU memory usage (in GB). We re-time these models on our hardware (a 3090 GPU) for a fair comparison. We report the $\mathcal{J}@60s$ with different interaction steps. $\mathcal{J}@60s$-$ith$ denotes the performance with up to $i$ human interactions.}
        \label{tab:performance_ivos_before}
        \centering
        \setlength{\tabcolsep}{1.4mm}
        \input{./tables/performance_ivos_before}
        \vspace{7mm}
    \end{table*}

\section{Experiments}
    Upon the construction of the LVOS dataset, some questions naturally arise:
    \begin{itemize}
      \item {\textbf{Question 1:} What happens when existing video object segmentation models, fitted for short-term video domains, come across long-term videos?}
      \item {\textbf{Question 2:} Which factors are instrumental in contributing to the variance in performance?}
      \item {\textbf{Question 3:} How to equip a video object segmentation model with the ability to handle long-term videos? }
      
    \end{itemize}
    In response to these inquiries, we have designed a series of experiments to provide answers and perform an analysis of the long-term video object segmentation task.

	\subsection{Experiment Setup}
	\textbf{Experiment Settings.}
    We carry out experiments encompassing four representative video object segmentation tasks, namely, semi-supervised video object segmentation, unsupervised single video object segmentation, unsupervised multiple video object segmentation, and interactive video object segmentation. The 720 videos are divided into 420 for training, 140 for validation, and the remaining 160 serve as test videos. For each object, we provide mask or scribbles as a reference.

    \textbf{Evaluation Metrics.} We adopt the two commonly used evaluation metrics, region similarity $\mathcal{J}$ and contour accuracy $\mathcal{F}$ as metrics, following DAVIS~\cite{perazzi2016benchmark,perazzi2017learning} and YouTube-VOS~\cite{xu2018youtube}. The region similarity $\mathcal{J}$ calculates the Intersection-over-Union(IOU) between groundtruth $G$ and prediction $M$, which is defined as $\mathcal{J} = \frac{M\cap G}{M\cup G}$. The contour accuracy $\mathcal{F}$ evaluates the precision of the segmentation boundary as the harmonic mean of the contour recall $P_{c}$ and contour precision $R_{c}$, which is defined as $\mathcal{F} = \frac{2P_{c}R_{c}}{P_{c}+R_{c}}$. Following YouTube-VOS~\cite{xu2018youtube}, we separately report the performance of seen categories and unseen categories, and obtain the final scores $\mathcal{J} \& \mathcal{F} = (\mathcal{J}_{s} + \mathcal{F}_{s} + \mathcal{J}_{u} + \mathcal{F}_{u})/4$, where subscript $s$ and $u$ denote scores in seen and unseen categories. 
    Through the separate evaluations of seen and unseen categories, we can  better assess the generalization ability of video object segmentation models. 
    
    \begin{table*}[htbp]
    	\centering        
        \caption{Attribute-based aggregate performance on validation sets. For each method, the column on the left represents the $\mathcal{J} \& \mathcal{F}$ over all sequences possessing that specific attribute (e.g., BC). Conversely, the right column indicates the performance gain (or loss) for that method for the remaining sequences without that respective attribute.}
    	\label{tab:attribute} 
        \setlength{\tabcolsep}{2.55mm}
        \input{./tables/performance_attribute_new}
    	\scriptsize

    \end{table*}

    \subsection{Domain Transfer Results (Question 1)}
    \label{sec:performance_before}
    We assess the performance of existing VOS models, which are only trained on short-term video datasets, on our LVOS validation and test sets. These evaluations encompass four representative tasks of video object segmentation.

    \textbf{Semi-supervised Video Object Segmentation.} Semi-supervised video object segmentation models rely on the first frame mask, establishing it as a conventional and widely adopted task within the field of video object segmentation. We assess 8 models specifically designed semi-supervised VOS models along with 2 visual foundation models, varying in terms of their design of matching mechanism and memory feature construction. We conduct experiments on these models with the weights pretrained on short-term video datasets, and the results are shown in Table~\ref{tab:performance_semi_before}. During the evaluation process, we restrict the memory length to 6 when assessing approaches with memory bank, such as STCN~\cite{cheng2021rethinking} and DeAOT~\cite{yang2022decoupling}. For a fair comparison, all the videos are down-sampled to 480p resolution. Despite the promising performance on short-term video datasets, these models illustrated a profound performance decline (about 25$\% \mathcal{J} \& \mathcal{F}$ decrease) when applied on LVOS. LWL~\cite{bhat2020learning} introduced a few-shot learner based on the appearance in the first frame. However, the initial frame's appearance may considerably differ from intermediate frames in a long-term video, resulting in LWL attaining 60.6$\%\mathcal{J} \& \mathcal{F}$ on validation sets and 60.9$\%\mathcal{J} \& \mathcal{F}$ on test sets. CFBI~\cite{yang2020collaborative} utilized L2 distance to calculate feature similarity. However, this approach proves unstable when there exist background objects similar with target object. Given the frequent background confusion in LVOS, CFBI's performance is unsatisfied. AFB-URR~\cite{liang2020video} and RDE~\cite{li2022recurrent} compressed the historical information into a fixed-size memory feature which inadvertently leads to information loss, thereby attributing to these models' poor accuracy on LVOS. DeAOT-B~\cite{yang2022decoupling} only relied on the previous and first frame, failing to fully exploit the temporal information, a crucial element for long-term video tasks. DeAOT-L~\cite{yang2022decoupling}, STCN~\cite{cheng2021rethinking}, and XMem~\cite{cheng2022xmem}, although capable of leveraging additional information through the memory bank, are still significantly constrained due to memory limitations. The small temporal windows fail to provide adequate historical information necessary for the segmentation of target object in current frame, resulting in the performance degradation on LVOS. The performance of visual foundation models such as SAM-PT~\cite{rajivc2023segment} and SegGPT~\cite{wang2023seggpt} on LVOS is worse and far from satisfactory. SAM-PT~\cite{rajivc2023segment}, built upon point tracking models~\cite{karaev2023cotracker,sand2008particle}, suffers from poor performance due to the inferior success of these point tracking models on long-term videos, consequently leading to subsequent segmentation errors. The poor performance of SegGPT~\cite{wang2023seggpt} also demonstrates that existing visual foundation models can not handle the complex motion in long-term videos.
    
    \textbf{Unsupervised Video Single Object Segmentation.} The objective of unsupervised video single object segmentation is to identify and segment the salient objects within the video completely autonomously, without any human intervention. 
    We evaluate 6 models specifically designed unsupervised video single object segmentation models. Following~\cite{hong2023simulflow,pei2022hierarchical}, we reshape the image into the shape $512 \times 512$, and generate the optical flow by using RAFT~\cite{teed2020raft}. Because LVOS is multiple object datasets, we choose 73 videos from validation and test sets wherein only a single salient object is annotated. The results are shown in Table~\ref{tab:performance_uvos_s_before}. These models only achieve roughly 40$\% \mathcal{J} \& \mathcal{F}$ on LVOS, a stark contrast to their performance on DAVIS-2016~\cite{perazzi2016benchmark}, where they reach about 90$\% \mathcal{J} \& \mathcal{F}$. These models largely depended on the motion between adjusted frames. Consequently, they tend to struggle with long-term videos due to complex motions and similar backgrounds, often leading them to segment all moving objects rather than the specific target object. Moreover, these models fail to detect the disappearance when the target object disappears, mistakenly identifying unrelated objects as the foreground. Due to the lack of long-term temporal information, the performance of existing unsupervised video single object segmentation models is far from satisfactory.

	\begin{figure*}[htbp]
		\centering
		\includegraphics[width=1.0\linewidth]{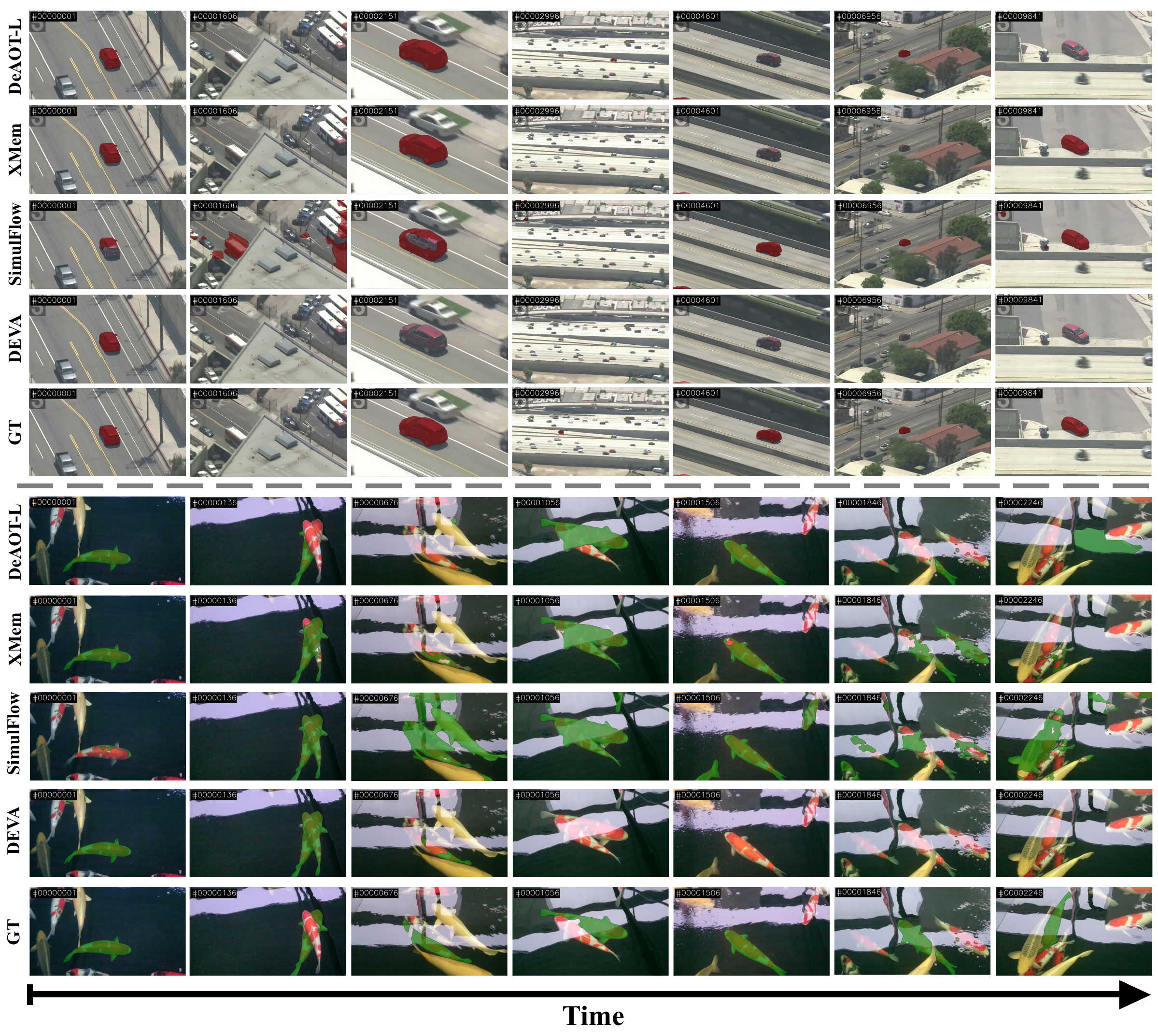}
		\caption{Visualization results of existing VOS models on LVOS. Best view in color.}
		\label{fig:vis_before}
	\end{figure*}

    \textbf{Unsupervised Video Multiple Object Segmentation}
    In the case of unsupervised video multiple object segmentation task, models are required to execute instance-level segmentation, thereby distinguishing and segmenting multiple objects within the same video. 
    We evaluate 2 models specifically designed unsupervised video multiple object segmentation models and an ensemble model. We report the results in Table~\ref{tab:performance_uvos_m_before}. We also observe the great performance drop on LVOS. For example, RVOS~\cite{ventura2019rvos} and STEm-Seg~\cite{athar2020stem} only achieve about 35 $\% \mathcal{J} \& \mathcal{F}$. DEVA~\cite{cheng2023tracking}, despite displaying promising results (over 60 $\% \mathcal{J} \& \mathcal{F}$), is dependent on the ensemble of multiple models and thus operates at a slower speed. Existing models generally built upon off-the-shelf instance segmentation methods, connecting detected objects in each frame in the temporal domain. While these instance segmentation methods are capable of detecting objects in each frame, the complicated scene change and frequent disappearance have significantly negative impact on the temporal linking. Furthermore, current models segment the video into several clips, which does not effectively utilize the available temporal information.

    \textbf{Interactive Video Object Segmentation} In the above three tasks, human interaction with VOS models is limited to a maximum of one instance, and the segmentation errors can not be subsequently corrected. For interactive VOS tasks, multiple interactions are permitted, allowing models to rectify segmentation inaccuracies via repeated interactions. We evaluate 2 interactive VOS models on LVOS, and report the performance for different numbers of interactions in Table~\ref{tab:performance_ivos_before}. Because the evaluation of contour accuracy $\mathcal{F}$ is time-consuming, we just assess the region similarity $\mathcal{J}$. It is observed that the performance of these models does improve as the number of interactions increases. However, there remains a considerable gap between performance on DAVIS 2019~\cite{caelles20192019} and LVOS.
    
    \begin{table*}[htbp]
    	\caption{Results of semi-supervised video object segmentation models on validation and test set after training on LVOS. Subscript $s$ and $u$ denote scores in seen and unseen categories. \textcolor{red}{$\uparrow$} represents the performance difference before and after training with LVOS training sets. Mem denotes the maximum GPU memory usage (in GB). We re-time these models on our hardware (a 3090 GPU) for a fair comparison.}
    	\label{tab:performance_semi_after}
    	\centering
    	\setlength{\tabcolsep}{3.3mm}
    	\input{./tables/performance_semi_after}
    \end{table*}
    
    \begin{table*}[htbp]
    	\caption{Results of unsupervised video single object segmentation models on validation and test set after training on LVOS. Subscript $s$ and $u$ denote scores in seen and unseen categories. \textcolor{red}{$\uparrow$} represents the performance difference before and after training with LVOS training sets. Mem denotes the maximum GPU memory usage (in GB). We re-time these models on our hardware (a 3090 GPU) for a fair comparison.}
    	\label{tab:performance_uvos_s_after}
    	\centering
    	\setlength{\tabcolsep}{3.3mm}
    	\input{./tables/performance_uvos_s_after}
    \end{table*}
    
    \begin{table*}[htbp]
    	\caption{Results of unsupervised video multiple object segmentation models on validation and test set after training on LVOS. Subscript $s$ and $u$ denote scores in seen and unseen categories. \textcolor{red}{$\uparrow$} represents the performance difference before and after training with LVOS training sets. Mem denotes the maximum GPU memory usage (in GB). We re-time these models on our hardware (a 3090 GPU) for a fair comparison.}
    	\label{tab:performance_uvos_m_after}
    	\centering
    	\setlength{\tabcolsep}{3.3mm}
    	\input{./tables/performance_uvos_m_after}
    \end{table*}

    \subsection{Attribute-based Evaluation (Question 2)}
    \label{sec:performance_attr}
    In Section~\ref{sec:performance_before}, we have conducted a series of experiments on LVOS in several settings and observed a great performance drop, while the reason of this decline is still unknown. Herein, we step further to explore the underlying reasons for this accuracy downgrade. We report the performance on subsets of the validation sets characterized by a particular attribute in Table~\ref{tab:attribute}. BC (Background Clutter), MB (Motion Blur), FM (Fast Motion), LR (Low Resolution), OCC (Occlusion), OV (Out-of-view), SV (Scale Variation) are all well-known challenges in short-term video segmentation, which have a great influence on long-term video performance. Furthermore, long-term videos introduce some unique challenges, such as LRA (Long-term Reappearance) and CTC (Cross-temporal Confusion), which also adversely affect performance. To further comprehend the reason of the significant performance degradation, we visualize the results of some models in Figure~\ref{fig:vis_before}. In the top video, the target car undergoes significant scale variations and frequent disappearance due to occlusion or going out-of-view, making the video challenging. In the frame 1606, unsupervised video single object segmentation models represented by SimulFlow struggled to identify the disappearance of the target object and segment the foreground content. In the frame 2151, after enduring a long-term occlusion, DEVA failed to redetect the car once it reappeared. In the frame 2996, the low resolution of the car and background interference led to the current VOS models either failing to detect the target objects or misrecognizing background objects as foreground entities. Due to the drastic change in the target car's appearance, some VOS models could not identify the target object in the latter half of the video. In the second video, the target goldfish's motion is complex, compounded by numerous similar background objects interacting extensively with the foreground object. Existing VOS models failed to identify and segment the foreground goldfish accurately.

    Based on the attribute-based performance analysis and visualization results, we propose that the significant degradation in performance is due to complex motion, large scale variations, frequent disappearances. Current VOS models are incapable of perceiving the disappearance and reappearance of objects, often mistakenly identifying background objects as target objects, and failing to re-identify the reappeared targets. In addition, the longer duration of the videos and similar background objects have a more negative impact on the detection and segmentation of target objects. The complexity of object motion could lead to severe occlusion or frequent disappearances, thereby demanding a stronger ability of VOS models. Challenges such as similar object, large scale variations, frequent disappearances, complex motion, and so on, are the primary reason for the significant performance decrease of existing VOS models on long-term video tasks. Some of these challenges are specific in long-term video tasks, while others are sharpened by the extended length of the videos.

    \begin{table*}[htbp]
        \setlength{\tabcolsep}{2.85mm} 
    	\centering
    	\caption{Oracle experiments on LVOS. \textcolor{red}{$\uparrow$} represents the performance improvement. }
    	\label{tab:oracle}
        \input{./tables/oracle}
    
    \end{table*}
    
    \subsection{Main Results on LVOS (Question 3)}
    \label{sec:performance_after}
    In Section~\ref{sec:performance_attr}, we have conducted an analysis to clarify the reason for the great performance degradation on LVOS based on the attribute-based evaluation and  visualization results. In this section, we design a series of experiments to explore how to empower a video object segmentation model with capability of handling long-term videos. 
    
    \textbf{Quantitative Evaluation.} We add LVOS training videos into the training data of previous VOS models and subsequently retrain them by following their origin training settings. We report the results of these models in Table~\ref{tab:performance_semi_after},~\ref{tab:performance_uvos_m_after}, and~\ref{tab:performance_uvos_s_after}. It is noteworthy that almost all models exhibited significant performance improvement on long-term videos after training on LVOS training sets. This enhancement demonstrates the potential of LVOS, with its rich scenarios and complex motion, to optimize the performance of video object segmentation models on long-term videos.

    \textbf{Oracle Experiments.} 
    To delve deeper into object localization and association, we carry out oracle experiments on DeAOT-B~\cite{yang2022decoupling}, DeAOT-L~\cite{yang2022decoupling}, and XMem~\cite{cheng2022xmem}. Results are shown in Table~\ref{tab:oracle}. In the row 2, we confine the object location within a box area determined by an oracle groundtruth mask. This leads to an average performance enhancement for these models of approximately 10$\% \mathcal{J} \& \mathcal{F}$, thereby confirming that segmentation errors result from poor tracking between similar objects. In the row 3, we replace the prediction with oracle groundtruth masks and store them into the memory feature bank. Upon resolving the segmentation errors, these models yield a higher score (about 20 $\% \mathcal{J} \& \mathcal{F}$ boost). This indicates that error accumulation is the primary cause of inaccuracies. However, in row 4, even when the accurate masks and locations are provided, there still exists a large gap between the result (up to 91.5 $\%\mathcal{J} \& \mathcal{F}$) and groundtruth. The gap signifies that complex movements are still very challenging for existing VOS models. In short, error accumulation is the main cause of unsatisfactory performances, and a more identifiable object representation is also important for distinguishing the target object from similar background. It is imperative for a robust VOS model to effectively accommodate the heightened complexity of motion in long-term videos.

    \subsection{Experiment Summary}
    In previous sections, we have undertaken extensive experiments to comprehensively analyze long-term video object segmentation tasks and attempt to investigate and address the three questions initially raised based on our proposed LVOS datasets. It has been found that existing VOS models, mainly designed for short-term videos, exhibit a marked performance decline when applied to long-term videos, as illustrated in Section~\ref{sec:performance_before}. To further this exploration, we implement an attribute-based evaluation and visualize some prediction results on LVOS in Section~\ref{sec:performance_attr}. We find that complex motion, large scale variations, frequent disappearances, and similar background are the main reason of the significant performance drop. These challenges, while also present in the shorter videos, become increasingly challenging with extended duration. In Section~\ref{sec:performance_after}, we explore the keys to enhancing the performance of current VOS models on long-term video tasks. The diverse scenarios and complex motion in LVOS videos can significantly boost the accuracy of these models. Additional oracle experiments also suggest that error accumulation is another critical factor contributing to poor performance. This problem is less pronounced in short-form video due to the shortness of the video, while it is very serious in longer duration videos. The experiments show that the performance degradation comes from the increase in video length, which further illustrates the necessity and significance of LVOS. For the design of VOS models,a robust and stronger object representation is necessary to distinguish the target object from similar background. Moreover, the temporal association mechanism and the ability to perceive the disappearance of target object are fundamental. Importantly, how to minimize the negative impact of error accumulation is also a very important issue. Additionally, the compression and utilization of historical information is a question worth exploring. We hope our experiments can provide insightful and potential directions for future VOS model design in real-world scenarios.

%% file: tables/performance_semi_before.tex
\begin{tabular}{l|c|c|c|ccccc|ccccc}
    \toprule
    \multirow{2}{*}{Methods}       & \multirow{2}{*}{FPS}       & \multirow{2}{*}{Mem}  & YTB  & \multicolumn{5}{c|}{Validation}   & \multicolumn{5}{c}{Test} \\
    \cline{5-14}
     & & & $\mathcal{J \& F}$ & $\mathcal{J \& F}$ & $\mathcal{J}_{s}$ &  $\mathcal{F}_{s}$ &  $\mathcal{J}_{u}$ &  $\mathcal{F}_{u}$ & $\mathcal{J \& F}$ & $\mathcal{J}_{s}$ &  $\mathcal{F}_{s}$ &  $\mathcal{J}_{u}$ &  $\mathcal{F}_{u}$ \\
            
    \midrule
    LWL~\cite{bhat2020learning}        &  4.5      & 1.93 & 81.5 & 60.6 \textcolor{blue}{$\downarrow_{+20.9}$} & 58.0  & 64.3  & 57.2    & 62.9    & 60.9 \textcolor{blue}{$\downarrow_{+20.6}$} & 54.5   & 59.5  & 62.1    & 67.4    \\
    AFB-URR~\cite{liang2020video}      &  2.7      & 5.08 & 79.6 & 44.1 \textcolor{blue}{$\downarrow_{+35.5}$} & 42.1  & 47.1  & 43.8    & 48.9    & 45.5 \textcolor{blue}{$\downarrow_{+34.1}$} & 42.1   & 47.1  & 43.8    & 48.9    \\
    CFBI~\cite{yang2020collaborative}  &  5.8      & 8.67 & 81.0 & 55.0 \textcolor{blue}{$\downarrow_{+26.0}$} & 52.9  & 59.2  & 51.7    & 56.2    & 53.6 \textcolor{blue}{$\downarrow_{+27.4}$}  & 53.0  & 57.0  & 50.4    & 54.2    \\
    DeAOT-B~\cite{yang2022decoupling}  &  38.9     & 3.76 & 84.6 & 63.3 \textcolor{blue}{$\downarrow_{+21.3}$} & 60.8  & 68.8  & 58.0    & 65.7    & 61.3 \textcolor{blue}{$\downarrow_{+23.3}$}  & 58.0  & 63.7  & 58.7    & 64.9    \\
    DeAOT-L~\cite{yang2022decoupling}  &  31.5     & 3.98 & 84.8 & 63.9 \textcolor{blue}{$\downarrow_{+20.9}$} & 61.5  & 69.0  & 58.4    & 66.6    & 63.8 \textcolor{blue}{$\downarrow_{+21.0}$} & 58.7   & 65.1  & 62.0    & 69.5    \\
    STCN~\cite{cheng2021rethinking}    &  28.9     & 1.45 & 83.0 & 60.6 \textcolor{blue}{$\downarrow_{+22.4}$} & 57.2  & 64.0  & 57.5    & 63.8    & 58.9 \textcolor{blue}{$\downarrow_{+24.1}$} & 55.9   & 61.0  & 56.6    & 62.0    \\
    RDE~\cite{li2022recurrent}         &  33.2     & 1.46 & 83.3 & 62.2 \textcolor{blue}{$\downarrow_{+21.1}$} & 56.7  & 64.1  & 60.8    & 67.2    & 58.1 \textcolor{blue}{$\downarrow_{+25.2}$} & 55.2   & 61.0  & 55.4    & 61.0    \\
    XMem~\cite{cheng2022xmem}          &  36.1     & 1.84 & 85.7 & 64.5 \textcolor{blue}{$\downarrow_{+21.2}$} & 62.6  & 69.1  & 60.6    & 65.6    & 63.9 \textcolor{blue}{$\downarrow_{+21.8}$} & 61.6   & 66.8  & 60.8    & 66.5    \\
    \midrule
    SAM-PT~\cite{rajivc2023segment}    &  1.5      & 2.81 & 76.2 & 15.9 \textcolor{blue}{$\downarrow_{+60.3}$} & 17.2  & 19.8  & 12.4    & 14.2    & 22.1 \textcolor{blue}{$\downarrow_{+54.1}$} & 22.3   & 25.5  & 19.1    & 21.6    \\
    SegGPT~\cite{wang2023seggpt}       &  0.8      & 6.43 & 74.7 & 30.7 \textcolor{blue}{$\downarrow_{+44.0}$} & 23.2  & 26.8  & 35.2    & 37.7    & 29.4 \textcolor{blue}{$\downarrow_{+45.4}$} & 24.3   & 26.6  & 33.1    & 33.4    \\
    \bottomrule
\end{tabular}

%% file: tables/performance_uvos_s_before.tex
\begin{tabular}{l|c|c|c|ccccc|ccccc}
    \toprule
    \multirow{2}{*}{Methods}       & \multirow{2}{*}{FPS}       & \multirow{2}{*}{Mem}  & DAVIS  & \multicolumn{5}{c|}{Validation}   & \multicolumn{5}{c}{Test} \\
    \cline{5-14}
     & & & $\mathcal{J \& F}$ & $\mathcal{J \& F}$ & $\mathcal{J}_{s}$ &  $\mathcal{F}_{s}$ &  $\mathcal{J}_{u}$ &  $\mathcal{F}_{u}$ & $\mathcal{J \& F}$ & $\mathcal{J}_{s}$ &  $\mathcal{F}_{s}$ &  $\mathcal{J}_{u}$ &  $\mathcal{F}_{u}$ \\
            
    \midrule
    AMC-Net~\cite{yang2021learning}       & 11.8  & 0.89  & 84.6     & 39.3 \textcolor{blue}{$\downarrow_{+45.3}$} & 40.3  & 47.7  & 29.5    & 39.5    & 44.6 \textcolor{blue}{$\downarrow_{+44.0}$} & 37.4  & 45.6  & 38.5     & 56.8    \\
    FSNet~\cite{ji2021full}               & 19.9  & 1.02  & 83.3     & 42.9 \textcolor{blue}{$\downarrow_{+40.4}$} & 44.0  & 51.0  & 33.1    & 43.7    & 45.8 \textcolor{blue}{$\downarrow_{+37.5}$} & 40.9  & 48.7  & 37.7     & 56.0    \\
    TMO~\cite{cho2023treating}            & 53.7  & 2.08  & 86.1     & 42.8 \textcolor{blue}{$\downarrow_{+43.3}$} & 47.9  & 55.2  & 28.6    & 39.7    & 45.9 \textcolor{blue}{$\downarrow_{+41.2}$} & 44.1  & 51.5  & 36.4     & 51.6    \\
    HFAN~\cite{pei2022hierarchical}       & 30.4  & 1.79  & 86.7     & 39.5 \textcolor{blue}{$\downarrow_{+47.2}$} & 45.5  & 50.8  & 25.3    & 36.4    & 42.7 \textcolor{blue}{$\downarrow_{+44.0}$} & 44.4  & 51.7  & 29.3     & 45.5    \\
    Isomer~\cite{yuan2023isomer}          & 24.6  & 0.61  & 90.0     & 37.7 \textcolor{blue}{$\downarrow_{+52.3}$} & 41.3  & 47.6  & 25.6    & 36.4    & 45.3 \textcolor{blue}{$\downarrow_{+44.7}$} & 40.7  & 47.8  & 36.4     & 56.2    \\
    SimulFlow~\cite{hong2023simulflow}    & 63.7  & 0.79  & 87.4     & 42.7 \textcolor{blue}{$\downarrow_{+44.7}$} & 50.4  & 55.8  & 25.6    & 39.1    & 44.1 \textcolor{blue}{$\downarrow_{+43.3}$} & 47.7  & 53.3  & 29.3     & 45.9    \\
    \bottomrule
\end{tabular}

%% file: tables/performance_uvos_m_before.tex
\begin{tabular}{l|c|c|c|ccccc|ccccc}
    \toprule
    \multirow{2}{*}{Methods}       & \multirow{2}{*}{FPS}       & \multirow{2}{*}{Mem}  & DAVIS  & \multicolumn{5}{c|}{Validation}   & \multicolumn{5}{c}{Test} \\
    \cline{5-14}
     & & & $\mathcal{J \& F}$ & $\mathcal{J \& F}$ & $\mathcal{J}_{s}$ &  $\mathcal{F}_{s}$ &  $\mathcal{J}_{u}$ &  $\mathcal{F}_{u}$ & $\mathcal{J \& F}$ & $\mathcal{J}_{s}$ &  $\mathcal{F}_{s}$ &  $\mathcal{J}_{u}$ &  $\mathcal{F}_{u}$ \\
            
    \midrule
    RVOS~\cite{ventura2019rvos}       & 22.1  & 4.72  & 43.7   & 28.8 \textcolor{blue}{$\downarrow_{+14.9}$} & 27.2  & 34.5  & 24.3    & 29.3    & 30.1 \textcolor{blue}{$\downarrow_{+13.6}$} & 31.6  & 36.5  & 24.0    & 28.2    \\
    STEm-Seg~\cite{athar2020stem}     & 18.9  & 5.52  & 64.7   & 38.1 \textcolor{blue}{$\downarrow_{+26.6}$} & 39.1  & 43.9  & 31.8    & 37.4    & 38.4 \textcolor{blue}{$\downarrow_{+26.3}$} & 42.5  & 45.7  & 31.5    & 33.9    \\
    DEVA~\cite{cheng2023tracking}     & 2.4   & 3.53  & 73.4   & 64.3 \textcolor{blue}{$\downarrow_{+9.10}$} & 59.8  & 67.9  & 60.1    & 69.2    & 62.0 \textcolor{blue}{$\downarrow_{+11.4}$} & 59.2  & 66.8  & 57.5    & 64.6   \\
    \bottomrule

\end{tabular}

%% file: tables/performance_ivos_before.tex
\begin{tabular}{l|c|c|c|cccc|cccc}
    \toprule
    \multirow{2}{*}{Methods}       & \multirow{2}{*}{FPS}       & \multirow{2}{*}{Mem}  & DAVIS  & \multicolumn{4}{c|}{Validation}   & \multicolumn{4}{c}{Test} \\
    \cline{5-12}
     & & & $\mathcal{J}@60s$ & $\mathcal{J}@60s$-1st & $\mathcal{J}@60s$-3rd &  $\mathcal{J}@60s$-5th &  $\mathcal{J}@60s$-8th &  $\mathcal{J}@60s$-1st & $\mathcal{J}@60s$-3rd &  $\mathcal{J}@60s$-5th &  $\mathcal{J}@60s$-8th  \\
            
    \midrule
    MiVOS~\cite{cheng2021modular}       & 7.2    & 12.75    & 85.4  & 6.5~\textcolor{blue}{$\downarrow_{+78.9}$} & 45.8~\textcolor{blue}{$\downarrow_{+39.6}$} & 49.0~\textcolor{blue}{$\downarrow_{+36.4}$} & 55.8~\textcolor{blue}{$\downarrow_{+29.6}$} & 9.6~\textcolor{blue}{$\downarrow_{+75.8}$}&  46.9~\textcolor{blue}{$\downarrow_{+38.5}$} & 49.1~\textcolor{blue}{$\downarrow_{+36.3}$} & 58.1~\textcolor{blue}{$\downarrow_{+27.3}$}  \\
    STCN~\cite{cheng2021rethinking}     & 15.4   & 13.51    & 85.5  & 6.5~\textcolor{blue}{$\downarrow_{+79.0}$} & 51.5~\textcolor{blue}{$\downarrow_{+34.0}$} & 54.2~\textcolor{blue}{$\downarrow_{+31.3}$} & 61.1~\textcolor{blue}{$\downarrow_{+24.4}$} & 9.4~\textcolor{blue}{$\downarrow_{+76.1}$}  &  48.3~\textcolor{blue}{$\downarrow_{+37.2}$} & 50.0~\textcolor{blue}{$\downarrow_{+35.5}$} & 59.5~\textcolor{blue}{$\downarrow_{+26.0}$}\\
    \bottomrule

\end{tabular}

%% file: tables/performance_attribute_new.tex
\begin{tabular}{l|llll|lll|lll}
\toprule
 \multirow{2}{*}{Attr} & \multicolumn{4}{c|}{Semi-supervised} & \multicolumn{3}{c|}{Unsupervised Multiple} & \multicolumn{3}{c}{Unsupervised Single} \\
 \cline{2-11}
 & DeAOT-B  & DeAOT-L  & RDE  & Xmem & RVOS & STEm-Seg & DEVA & HFAN & Isomer & SimulFlow \\
\midrule
BC                 & 56.4${}_{+6.0}$   & 56.0${}_{+7.6}$  & 50.9${}_{+8.3}$  & 52.1${}_{+10.6}$  & 26.2${}_{+7.7}$  & 32.2${}_{+10.2}$          & 59.8${}_{-2.5}$     & 45.5${}_{+3.3}$  & 44.2${}_{-3.0}$  & 48.7${}_{+6.4}$   \\
DEF                & 65.3${}_{-9.5}$ & 64.3${}_{-7.8}$ & 58.4${}_{-6.4}$ & 64.1${}_{-11.5}$ & 32.4${}_{-4.8}$ & 43.8${}_{-11.2}$         & 66.8${}_{-11.3}$ & 56.5${}_{-13.8}$ & 46.6${}_{-6.0}$  & 62.6${}_{-15.6}$ \\
MB                 & 53.0${}_{+8.3}$   & 54.7${}_{+6.1}$   & 48.1${}_{+8.6}$   & 51.6${}_{+6.5}$   & 28.8${}_{+0.5}$  & 32.2${}_{+5.6}$           & 54.2${}_{+7.1}$   & 48.9${}_{-2.1}$  & 42.7${}_{+0.1}$  & 53.6${}_{-1.9}$   \\
FM                 & 59.4${}_{-1.8}$  & 61.2${}_{-5.3}$  & 51.9${}_{+4.5}$   & 57.1${}_{-2.4}$  & 28.2${}_{+1.8}$ & 32.5${}_{+7.5}$           & 59.4${}_{-1.2}$  & 48.8${}_{-3.7}$  & 42.9${}_{-0.7}$  & 53.6${}_{-3.4}$  \\
LR                 & 53.5${}_{+11.1}$  & 54.3${}_{+10.0}$  & 48.6${}_{+11.7}$  & 47.9${}_{+17.8}$  & 24.7${}_{+9.6}$ & 29.0${}_{+15.2}$          & 51.1${}_{+17.1}$    & 38.7${}_{+13.7}$  & 39.3${}_{+5.4}$   & 40.7${}_{+18.3}$    \\
OCC                & 57.4${}_{+5.2}$   & 57.5${}_{+6.0}$   & 54.3${}_{-1.6}$  & 55.7${}_{+1.3}$    & 28.8${}_{+1.1}$  & 33.7${}_{+9.8}$           & 58.3${}_{+2.3}$  & 48.2${}_{-2.9}$  & 44.3${}_{-5.0}$    & 52.9${}_{-2.2}$  \\
OV                 & 56.2${}_{+4.1}$   & 55.8${}_{+5.2}$   & 52.6${}_{+2.3}$   & 56.8${}_{-1.5}$   & 29.0${}_{-0.2}$  & 35.3${}_{+1.1}$           & 54.6${}_{+7.3}$   & 42.4${}_{+7.3}$  & 40.4${}_{+3.3}$  & 44.1${}_{+12.1}$  \\
SV                 & 58.5${}_{+1.1}$    & 58.7${}_{+1.5}$  & 54.3${}_{-4.4}$   & 55.3${}_{+8.0}$   & 28.4${}_{+7.7}$  & 34.3${}_{+17.8}$          & 58.9${}_{+0.1}$   & 45.9${}_{+6.5}$   & 41.7${}_{+4.4}$   & 50.4${}_{+8.5}$  \\
DB                 & 56.9${}_{+6.2}$   & 59.6${}_{-2.8}$  & 55.4${}_{-5.4}$  & 56.4${}_{-1.7}$  & 31.2${}_{-7.8}$ & 38.4${}_{-9.2}$         & 60.1${}_{-4.7}$  & 50.0${}_{-2.7}$  & 45.1${}_{-1.9}$  & 55.6${}_{-2.5}$    \\
SC                 & 59.2${}_{-1.0}$    & 56.6${}_{+4.0}$   & 57.9${}_{-7.0}$  & 59.2${}_{-5.7}$  & 31.2${}_{-3.8}$ & 36.7${}_{-1.5}$          & 61.3${}_{-4.5}$  & 54.2${}_{-9.8}$ & 49.7${}_{-10.0}$ & 56.3${}_{-5.8}$  \\
AC                 & 60.2${}_{-5.2}$    & 61.1${}_{-7.3}$  & 56.5${}_{-8.1}$   & 58.9${}_{-9.4}$  & 30.9${}_{-5.8}$ & 37.8${}_{-6.0}$         & 61.4${}_{-7.9}$   & 48.2${}_{-3.5}$  & 43.1${}_{-1.7}$   & 53.9${}_{-6.8}$  \\
LRA                & 53.0${}_{+9.8}$  & 51.9${}_{+12.3}$  & 51.4${}_{+4.4}$   & 52.7${}_{+5.7}$   & 27.3${}_{+3.2}$  & 31.4${}_{+7.9}$           & 53.6${}_{+9.7}$   & 44.5${}_{+5.4}$   & 38.9${}_{+7.4}$     & 46.9${}_{+10.6}$  \\
CTC                & 56.6${}_{+5.0}$     & 56.5${}_{+5.9}$   & 50.4${}_{+9.1}$   & 53.5${}_{+6.4}$   & 25.8${}_{+8.5}$  & 32.3${}_{+9.3}$            & 60.1${}_{-3.0}$   & 44.2${}_{+5.3}$   & 42.7${}_{-0.1}$  & 46.7${}_{+9.3}$  \\
\bottomrule
\end{tabular} 

%% file: tables/performance_semi_after.tex
\begin{tabular}{l|c|c|ccccc|ccccc}
    \toprule
    \multirow{2}{*}{Methods}       & \multirow{2}{*}{FPS}       & \multirow{2}{*}{Mem}   & \multicolumn{5}{c|}{Validation}   & \multicolumn{5}{c}{Test} \\
    \cline{4-13}
     & & & $\mathcal{J \& F}$ & $\mathcal{J}_{s}$ &  $\mathcal{F}_{s}$ &  $\mathcal{J}_{u}$ &  $\mathcal{F}_{u}$ & $\mathcal{J \& F}$ & $\mathcal{J}_{s}$ &  $\mathcal{F}_{s}$ &  $\mathcal{J}_{u}$ &  $\mathcal{F}_{u}$ \\
            
    \midrule
    LWL~\cite{bhat2020learning}        &  4.5      & 1.93  & 61.4 \textcolor{red}{$\uparrow_{+0.8}$} & 58.7  & 64.9  & 58.5    & 63.5    & 59.8 \textcolor{red}{$\uparrow_{-1.1}$}  & 57.8    & 62.5  & 57.0    & 61.8    \\
    AFB-URR~\cite{liang2020video}      &  2.7      & 5.08  & 49.8 \textcolor{red}{$\uparrow_{+5.1}$} & 45.5  & 51.3  & 48.3    & 54.0    & 48.7 \textcolor{red}{$\uparrow_{+3.2}$}  & 47.8    & 52.2  & 45.5    & 49.4    \\
    CFBI~\cite{yang2020collaborative}  &  5.8      & 8.67  & 54.6 \textcolor{red}{$\uparrow_{-0.4}$} & 51.9  & 59.0  & 50.7    & 56.7    & 58.0 \textcolor{red}{$\uparrow_{+4.4}$}  & 53.5  & 58.5  & 57.5    & 62.5    \\
    DeAOT-B~\cite{yang2022decoupling}  &  38.9     & 3.76  & 67.3 \textcolor{red}{$\uparrow_{+4.0}$} & 62.3  & 70.0  & 64.5    & 72.5    & 62.8  \textcolor{red}{$\uparrow_{+1.5}$}  & 60.6  & 66.2  & 59.3    & 65.3    \\
    DeAOT-L~\cite{yang2022decoupling}  &  31.5     & 3.98  & 67.1 \textcolor{red}{$\uparrow_{+3.2}$} & 64.8  & 72.4  & 61.7    & 69.3    & 65.9 \textcolor{red}{$\uparrow_{+2.1}$}  & 60.7  & 67.0  & 64.2    & 71.7    \\
    STCN~\cite{cheng2021rethinking}    &  28.9     & 1.45  & 62.3 \textcolor{red}{$\uparrow_{+1.7}$} & 59.2  & 66.3  & 59.3    & 64.5    & 60.7 \textcolor{red}{$\uparrow_{+1.8}$}  & 59.0  & 64.2  & 57.3    & 62.5    \\
    RDE~\cite{li2022recurrent}         &  33.2     & 1.46  & 65.1 \textcolor{red}{$\uparrow_{+2.9}$} & 59.0  & 66.2  & 63.7    & 71.7    & 62.1 \textcolor{red}{$\uparrow_{+4.0}$} & 59.3  & 65.4  & 58.9    & 64.8    \\
    XMem~\cite{cheng2022xmem}          &  36.1     & 1.84  & 67.4 \textcolor{red}{$\uparrow_{+2.9}$} & 65.3  & 72.4  & 62.9    & 68.8    & 65.2 \textcolor{red}{$\uparrow_{+1.3}$} & 62.4  & 67.8  & 62.0    & 68.6    \\
    \bottomrule
\end{tabular}

%% file: tables/performance_uvos_s_after.tex
\begin{tabular}{l|c|c|ccccc|ccccc}
    \toprule
    \multirow{2}{*}{Methods}       & \multirow{2}{*}{FPS}       & \multirow{2}{*}{Mem}   & \multicolumn{5}{c|}{Validation}   & \multicolumn{5}{c}{Test} \\
    \cline{4-13}
     & & & $\mathcal{J \& F}$ & $\mathcal{J}_{s}$ &  $\mathcal{F}_{s}$ &  $\mathcal{J}_{u}$ &  $\mathcal{F}_{u}$ & $\mathcal{J \& F}$ & $\mathcal{J}_{s}$ &  $\mathcal{F}_{s}$ &  $\mathcal{J}_{u}$ &  $\mathcal{F}_{u}$ \\
            
    \midrule
    AMC-Net~\cite{yang2021learning}       & 11.8  & 0.89  & 39.9 \textcolor{red}{$\uparrow_{+0.6}$} & 41.5  & 48.6  & 29.6    & 39.9    & 44.1 \textcolor{red}{$\uparrow_{-0.5}$} & 38.2  & 46.1  & 36.9    & 55.4    \\
    FSNet~\cite{ji2021full}               & 19.9  & 1.02  & 46.3 \textcolor{red}{$\uparrow_{+3.4}$} & 52.4  & 60.0  & 31.9    & 40.9    & 51.0 \textcolor{red}{$\uparrow_{+5.2}$} & 52.2  & 59.4  & 35.9    & 56.5    \\
    TMO~\cite{cho2023treating}            & 53.7  & 2.08  & 46.8 \textcolor{red}{$\uparrow_{+4.0}$} & 56.8  & 62.9  & 26.5    & 40.8    & 52.8 \textcolor{red}{$\uparrow_{+6.9}$} & 56.5  & 61.9  & 39.1    & 53.7    \\
    HFAN~\cite{pei2022hierarchical}       & 30.4  & 1.79  & 40.8 \textcolor{red}{$\uparrow_{+1.3}$} & 49.8  & 54.8  & 24.4    & 34.4    & 49.1 \textcolor{red}{$\uparrow_{+6.4}$} & 49.6  & 55.5  & 37.8    & 53.3    \\
    Isomer~\cite{yuan2023isomer}          & 24.6  & 0.61  & 42.9 \textcolor{red}{$\uparrow_{+5.2}$} & 51.0  & 57.0  & 26.8    & 36.7    & 49.6 \textcolor{red}{$\uparrow_{+4.3}$} & 53.2  & 58.9  & 32.6    & 53.7    \\
    SimulFlow~\cite{hong2023simulflow}    & 63.7  & 0.79  & 42.2 \textcolor{red}{$\uparrow_{-0.5}$} & 51.0  & 57.1  & 25.1    & 35.7    & 43.8 \textcolor{red}{$\uparrow_{-0.3}$} & 50.3  & 56.5  & 25.8    & 42.7    \\
    \bottomrule
\end{tabular}

%% file: tables/performance_uvos_m_after.tex
\begin{tabular}{l|c|c|ccccc|ccccc}
    \toprule
    \multirow{2}{*}{Methods}       & \multirow{2}{*}{FPS}       & \multirow{2}{*}{Mem}  & \multicolumn{5}{c|}{Validation}   & \multicolumn{5}{c}{Test} \\
    \cline{4-13}
     & &  & $\mathcal{J \& F}$ & $\mathcal{J}_{s}$ &  $\mathcal{F}_{s}$ &  $\mathcal{J}_{u}$ &  $\mathcal{F}_{u}$ & $\mathcal{J \& F}$ & $\mathcal{J}_{s}$ &  $\mathcal{F}_{s}$ &  $\mathcal{J}_{u}$ &  $\mathcal{F}_{u}$ \\
            
    \midrule
    RVOS~\cite{ventura2019rvos}       & 22.1  & 4.72  & 29.3 \textcolor{red}{$\uparrow_{+0.5}$} & 30.7  & 36.4  & 21.0    & 29.1    & 30.5 \textcolor{red}{$\uparrow_{+0.4}$} & 30.0  & 36.3  & 26.6    & 28.9    \\
    STEm-Seg~\cite{athar2020stem}     & 18.9  & 5.52  & 39.3 \textcolor{red}{$\uparrow_{+1.2}$} & 39.8  & 44.6  & 33.9    & 38.9    & 39.6 \textcolor{red}{$\uparrow_{+1.2}$} & 44.5  & 47.9  & 31.3    & 34.8    \\
    \bottomrule

\end{tabular}

%% file: tables/oracle.tex
\begin{tabular}{ccc|ccc|ccc}
\toprule
\multirow{3}{*}{\#} & \multirow{3}{*}{Oracle Box}  & \multirow{3}{*}{Oracle Mask}  & \multicolumn{3}{c|}{Validation}   & \multicolumn{3}{c}{Test} \\
\cline{4-9}
&             &             & DeAOT-B~\cite{yang2022decoupling}    & DeAOT-L~\cite{yang2022decoupling} & XMem~\cite{cheng2022xmem} & DeAOT-B~\cite{yang2022decoupling} & DeAOT-L~\cite{yang2022decoupling} & XMem~\cite{cheng2022xmem} \\
& &   & $\mathcal{J} \& \mathcal{F}$         & $\mathcal{J} \& \mathcal{F}$      & $\mathcal{J} \& \mathcal{F}$   & $\mathcal{J} \& \mathcal{F}$     & $\mathcal{J} \& \mathcal{F}$      & $\mathcal{J} \& \mathcal{F}$   \\
 \midrule
1 &             &                      & 67.3       & 67.1    & 73.3 & 62.8    & 65.9    & 70.8 \\
2 & \checkmark            &             & 73.8\textcolor{red}{$\uparrow_{+6.5}$}\phantom{1}       & 78.4\textcolor{red}{$\uparrow_{+11.3}$}    & 68.5\textcolor{red}{$\uparrow_{+4.8}$}\phantom{1}  & 72.1\textcolor{red}{$\uparrow_{+9.3}$}\phantom{1}    & 77.0\textcolor{red}{$\uparrow_{+11.1}$}    & 71.1\textcolor{red}{$\uparrow_{+0.3}$}\phantom{1} \\
2 &             & \checkmark           & 86.8\textcolor{red}{$\uparrow_{+19.5}$}       & 89.4\textcolor{red}{$\uparrow_{+22.3}$}    & 88.7\textcolor{red}{$\uparrow_{+15.4}$} & 85.6\textcolor{red}{$\uparrow_{+22.8}$}    & 88.8\textcolor{red}{$\uparrow_{+22.9}$}    & 88.9\textcolor{red}{$\uparrow_{+18.1}$} \\
3 & \checkmark            & \checkmark  & 88.8\textcolor{red}{$\uparrow_{+21.5}$}       & 90.3\textcolor{red}{$\uparrow_{+23.2}$}    & 89.1\textcolor{red}{$\uparrow_{+15.8}$}  & 89.4\textcolor{red}{$\uparrow_{+26.6}$}    & 91.5\textcolor{red}{$\uparrow_{+25.6}$}    & 89.5\textcolor{red}{$\uparrow_{+18.7}$} \\
\bottomrule
\end{tabular} 

%% file: section/5_future_and_conclusion.tex
\section{Future Works}
In this section, we discuss the challenges posed by our LVOS and provide some potential future research directions for long-term video object segmentation.
\begin{itemize}
      \item {\textbf{Long-term Dependence.} Addressing the complexities of object movements, occlusions, and shape deformations in video sequences necessitates models to adeptly interpret long-term dependencies for accurate object segmentation. However, current models often struggle with managing these dependencies, potentially leading to information loss, ineffective information correlation over time, or failure to adapt to complex object changes. These issues might impair tracking accuracy and segmentation performance. Hence, future research needs to enhance models' effectiveness in handling long-term dependencies.}
          
      \item {\textbf{Dynamic Scenes.} Long-term videos, comprising multiple scenes with unique lighting, backgrounds, and elements, challenge models' ability to adapt for accurate target segmentation. Models are tested for diversity management and dynamic responsiveness across varying scenes. Enhancing anti-interference, generalization, and dynamic adaptability capabilities should be a focal point of future research, alongside the incorporation of long-term memory mechanisms to improve target identification and tracking amid scene transitions.}
          
      \item {\textbf{Occluded or Disappeared Object.} Occlusion or disappearance of objects presents significant challenges to VOS models, which depend on frame-to-frame similarity. Models may lose track of an object when it becomes occluded or disappears, and struggle to reidentify and track it upon reappearance due to potential changes in attributes like shape, color, or position. As most models presume short-term attribute consistency, performance may falter in long-duration videos. Future models should consider more flexible and resilient feature representation methods to maintain stable tracking, even with significant changes in object attributes. }

      \item {\textbf{Stronger Spatial-Temporal Association.} Strengthening spatial-temporal association is crucial for long-term video target segmentation. This association encapsulates the object's spatial location and its temporal movements, enabling accurate and coherent target recognition and tracking. Future work could focus on refining the model's temporal and spatial aspects. Incorporation of Recurrent Neural Networks (RNN) or 3D Convolutional Neural Networks could improve the model's ability to capture temporal and spatial data. Additionally, advanced target tracking algorithms could enhance target tracking accuracy. Combining semantic and instance segmentation techniques could also be considered, aiding in target identification as well as recognizing the target's specific shape and location. }
          
      \item {\textbf{Memory Management.} Long-term videos, due to their length, demand considerably more memory than short-term videos, posing an obstacle as memory requirements scale with video length. Despite hardware advancements, memory is a finite resource that must be effectively managed and optimized for video object segmentation tasks. This is particularly salient for edge devices like surveillance cameras and mobile devices, where lower computing power and memory constraints present significant challenges for processing long-term video object segmentation. Future work should consider performance optimization under these constrained conditions.}

      \item {\textbf{Annotations Reliance.} Minimizing annotation reliance is a key focus for long-term video object segmentation. The complexity of this task has traditionally required extensive manual annotation, a process that is labor-intensive, time-consuming, and potentially error-prone. Hence, research aimed at reducing annotation dependence holds practical and theoretical value. Future studies could explore unsupervised or semi-supervised learning methods, enabling the model to lessen its reliance on detailed annotations and enhance its adaptability and robustness through self-learning.}

\end{itemize}

\section{Conclusion}
	In this paper, we introduce LVOS, a novel dataset specifically designed for long-term video object segmentation. Different from prevalent short-term VOS datasets, the mean duration of the videos in LVOS extends to 1.14 minutes. This increased complexity in terms of motion and duration imposes additional challenges to existing VOS models. We conduct comprehensive assessment of 20 existing VOS approaches across 4 different settings on LVOS. Through the attribute-based evaluation and result visualization, we analyze the main cause of the significant performance decline in long-term videos. The experiments demonstrate that the extension in video length is the direct reason for accuracy degradation, thus underscoring the necessity and importance of LVOS. We explore the key factor to enhance the precision of VOS models when deployed on long-term videos, providing potential directions for future research on long-term VOS. By presenting LVOS, we hope to provide a platform to encourage a comprehensive study of video object segmentation in real-world scenarios.